\documentclass[12pt,a4paper]{scrartcl}
\usepackage[utf8]{inputenc}
\usepackage[T1]{fontenc}
\usepackage{lmodern}
\usepackage[english]{babel} 
\usepackage{csquotes}
\usepackage[intlimits]{amsmath}
\usepackage{amssymb}
\usepackage{dsfont} 
\usepackage[pdftex]{graphicx}
\usepackage{microtype}
\usepackage[pdfborder={0 0 0}]{hyperref}
\usepackage{xifthen}
\usepackage{tikz}
\usepackage{verbatim}
\usepackage{hyperref}
\usetikzlibrary{automata,arrows,positioning,calc}

\usepackage{amsthm}

\makeatletter
\newcommand{\sectionauthor}[1]{%
  {\parindent0pt\vspace*{-5pt}%
  \linespread{1.1}\small\phantom{MM}#1%
  \par\nobreak\vspace*{35pt}}
  \@afterheading%
}
\makeatother

\usepackage{xspace}

\newcommand{\ie}{i.e.,\ }
\newcommand{\eg}{e.g.,\ }
\newcommand{\wrt}{w.r.t.\xspace}
\newcommand{\cf}{cf.\xspace}
\newcommand{\etal}{\xspace et al.\xspace}

\newcommand{\refsec}[1]{Sec.~\ref{#1}}

\usepackage{pdflscape,longtable} 

\usepackage[nopostdot,nogroupskip,style=super,nonumberlist,toc,automake,toc]{glossaries}

\usepackage{authblk}

\makeglossaries

\newglossaryentry{Nn}{name={NN},description={Neural Network}}
\newglossaryentry{Dnn}{name={DNN},description={Deep Neural Network}}
\newglossaryentry{Crn}{name={CRN},description={Conditional Random Fields}}
\newglossaryentry{Lstm}{name={LSTM},description={Long Short Term Memory}}
\newglossaryentry{Rnn}{name={RNN},description={Recurrent Neural Network}}

\title{Inspect, Understand, Overcome:\\A Survey of Practical Methods\\for AI Safety}
\author[1]{Sebastian Houben}
\author[2]{Stephanie Abrecht}
\author[1]{Maram Akila}
\author[15]{Andreas Bär}
\author[10]{Felix Brockherde}
\author[8]{Patrick Feifel}
\author[15]{Tim Fingscheidt}
\author[1]{Sujan Sai Gannamaneni}
\author[8]{Seyed Eghbal Ghobadi}
\author[8]{Ahmed Hammam}
\author[9]{Anselm Haselhoff}
\author[11]{Felix Hauser}
\author[2]{Christian Heinzemann}
\author[16]{Marco Hoffmann}
\author[7]{Nikhil Kapoor}
\author[13]{Falk Kappel}
\author[15]{Marvin Klingner}
\author[9]{Jan Kronenberger}
\author[9]{Fabian Küppers}
\author[15]{Jonas Löhdefink}
\author[16]{Michael Mlynarski}
\author[1]{Michael Mock}
\author[13]{Firas Mualla}
\author[14]{Svetlana Pavlitskaya}
\author[1]{Maximilian Poretschkin}
\author[16]{Alexander Pohl}
\author[4]{Varun Ravi-Kumar}
\author[1]{Julia Rosenzweig}
\author[5]{Matthias Rottmann}
\author[1]{Stefan Rüping}
\author[4]{Timo Sämann}
\author[7]{Jan David Schneider}
\author[1]{Elena Schulz}
\author[3]{Gesina Schwalbe}
\author[1]{Joachim Sicking}
\author[12]{Toshika Srivastava}
\author[7]{Serin Varghese}
\author[14]{Michael Weber}
\author[6]{Sebastian Wirkert}
\author[1]{Tim Wirtz}
\author[2]{Matthias Woehrle}

\affil[1]{Fraunhofer Institute for Intelligent Analysis and Information Systems}
\affil[2]{Robert Bosch GmbH}
\affil[3]{Continental AG}
\affil[4]{Valeo S.A.}
\affil[5]{University of Wuppertal}
\affil[6]{Bayerische Motorenwerke AG}
\affil[7]{Volkswagen AG}
\affil[8]{Opel Automobile GmbH}
\affil[9]{Hochschule Ruhr West}
\affil[10]{umlaut AG}
\affil[11]{Karlsruhe Institute of Technology}
\affil[12]{Audi AG}
\affil[13]{ZF Friedrichshafen AG}
\affil[14]{FZI Research Center for Information Technology}
\affil[15]{Technische Universität Braunschweig}
\affil[16]{QualityMinds GmbH}

\date{}

\begin{document}
\maketitle

\pagebreak

\begin{abstract}
The use of deep neural networks (DNNs) in safety-critical applications like mobile health and autonomous driving is challenging due to numerous model-inherent shortcomings. 
These shortcomings are diverse and range from a lack of generalization over insufficient interpretability to problems with malicious inputs. Cyber-physical systems employing DNNs are therefore likely to suffer from \emph{safety concerns}. 
In recent years, a zoo of state-of-the-art techniques aiming to address these safety concerns has emerged.
This work provides a structured and broad overview of them. 
We first identify categories of insufficiencies to then describe research activities aiming at their detection, quantification, or mitigation. 
Our paper addresses both machine learning experts and safety engineers: 
The former ones might profit from the broad range of machine learning (ML) topics covered and discussions on limitations of recent methods. 
The latter ones might gain insights into the specifics of modern ML methods. 
We moreover hope that our contribution fuels discussions on desiderata for ML systems and strategies on how to propel existing approaches accordingly.
\end{abstract}

\pagebreak
\tableofcontents
\pagebreak

\section{Introduction} \label{sec:intro} 
\sectionauthor{Sebastian Houben\textsuperscript{1}, Michael Mock\textsuperscript{1}, Timo Sämann\textsuperscript{4}, Gesina Schwalbe\textsuperscript{3}, Joachim Sicking\textsuperscript{1}}
In barely a decade, deep neural networks (DNNs) have revolutionized the field of machine learning by reaching unprecedented, sometimes superhuman, performances on a growing variety of tasks.
Many of these neural models have found their way into consumer applications like smart speakers, machine translation engines or content feeds. However, in safety-critical systems, where human life might be at risk, the use of recent DNNs is challenging as various model-immanent insufficiencies are yet difficult to address. 
\par
This paper summarizes the promising lines of research in how to identify, address, and at least partly mitigate these DNN insufficiencies.
While some of the reviewed works are theoretically grounded and foster the overall understanding of training and predictive power of DNNs, others provide practical tools to adapt their development, training or predictions. 
We refer to any such method as a \emph{safety mechanism} if it addresses one or several safety concerns in a feasible manner.
Their effectiveness in mitigating safety concerns is assessed by \emph{safety metrics}  \cite{odena_tensorfuzz_2019,cheng_towards_2018,schwalbe_survey_2020,burton_confidence_2019}. 
As most safety mechanisms target only a particular insufficiency, we conclude that a \emph{holistic safety argumentation}\cite{saemann_strategy_2020,schwalbe_survey_2020,burton_confidence_2019,willers2020safety} for a complex DNN-based systems will in many cases rely on a variety of safety mechanisms. 
\par
We structure our review of these mechanisms as follows: Chapter \ref{sec:dataset_optimization} focuses on \emph{dataset optimization} for network training and evaluation. It is motivated by the well-known fact that, in comparison to humans, DNNs perform poorly on data that is structurally different from training data. Apart from insufficient generalization capabilities of these models, the data acquisition process and distributional data shifts over time play vital roles. We survey potential counter-measures, \eg augmentation strategies and outlier detection techniques.
\par
Mechanisms that improve on \emph{robustness} are described in Chapters \ref{sec:robust_training} and \ref{sec:adversarial_attacks}, respectively. They deserve attention as DNNs are generally not resilient to common perturbations and adversarial attacks.
\par
Chapter \ref{sec:interpretability} addresses incomprehensible network behavior and reviews mechanisms that aim at \emph{explainability}, \eg a more transparent functioning of DNNs.
\par
Moreover, DNNs tend to overestimate their prediction confidence, especially on unseen data. 
Straightforward ways to estimate prediction confidence yield mostly unsatisfying results.
Among others, this observation fuelled research on more sophisticated \emph{uncertainty estimations} (see Chapter \ref{sec:uncertainty}), \emph{redundancy mechanisms} (see Chapter \ref{sec:aggregation}) and attempts to reach \emph{formal verification} as addressed in Chapter \ref{sec:verification}.

\par
At last, many safety-critical applications require not only accurate but also near real-time decisions. This is covered by mechanisms on the DNN \emph{architectural level}  (see Chapter \ref{sec:architecture}) and furthermore by \emph{compression} and \emph{quantization} methods (see Chapter \ref{sec:compression}).
\par

We conclude this review of mechanism categories with an outlook on the steps to transfer a carefully arranged combination of safety mechanisms into an actual holistic safety argumentation.

\section{Dataset Optimization}\label{sec:dataset_optimization}
\sectionauthor{Matthias Rottmann\textsuperscript{5}}
The performance of a trained model inherently relies on the nature of the underlying dataset.
For instance, a dataset with poor variability will hardly result in a model ready for real-world applications.
In order to approach the latter, data selection processes such as corner case selection and active learning are of utmost importance.
These approaches can help to design datasets that contain the most important information, while preventing the so much desired information from getting lost in an ocean of data.
For a given dataset and active learning setups, data augmentation techniques are very common aiming at extracting as much model performance out of the dataset as possible.

On the other hand, safety arguments also require the analysis of how a model behaves on out-of-distribution data, data that contains concepts the model has not encountered during training.
This is quite likely to happen as our world is under constant change, in other words exposed to a constantly growing domain shift.
Therefore, these fields are lately gaining interest, also with respect to perception in automated driving.

\subsection{Outlier/Anomaly Detection}\label{subsec:dataset_optimization:anomaly_detection}
\sectionauthor{Sujan Sai Gannamaneni\textsuperscript{1}, Matthias Rottmann\textsuperscript{5}}
The terms anomaly, outlier and \emph{out-of-distribution} (OOD) data detection are often used interchangeably in literature and refer to task of identifying data samples that are not representative of training data distribution. Uncertainty evaluation (\cf Chapter \ref{sec:uncertainty}) is closely tied to this field as self-evaluation of models is one of the active areas of research for OOD detection. In particular, for image classification problems it has been reported that neural networks often produce high confidence predictions on OOD data \cite{Nguyen2015, hendrycks2017baseline}. The detection of such OOD inputs can either be tackled by post-processing techniques that adjust the estimated confidence \cite{liang2018, devries2018} or by enforcing low confidence on OOD samples during training \cite{Hein2019, hendrycks2019oe}. Even guarantees that neural networks produce low confidence predictions for OOD samples can be provided under specific assumptions (\cf \cite{Meinke2020}). More precisely, this work utilizes Gaussian mixture models that, however, may suffer from high-dimensional data and require strong assumptions on the distribution parameters.
Some approaches use generative models like \emph{GANs} \cite{schlegl2017unsupervised, akcay2018ganomaly} and \emph{autoencoders} \cite{zhou2017anomaly} for outlier detection. The models are trained to learn in-distribution data manifolds and will produce higher reconstruction loss for outliers.
\par
For OOD detection in semantic segmentation, only a few works have been presented so far.
Angus \etal \cite{Angus2019} present a comparative study of common OOD detection methods, which mostly deal with image-level classification. In addition, they provide a novel setup of relevant OOD datasets for this task. Another work trains a fully convolutional binary classifier that distinguishes image patches from a known set of classes from image patches stemming from an unknown class \cite{Bevandic2018}. The classifier output applied at every pixel will give the per-pixel confidence value for an OOD object. Both of these works perform at pixel level and without any sophisticated feature generation methods specifically tailored for the detection of entire OOD instances.
Up to now, outlier detection has not been  studied extensively for object detection tasks based on benchmark object detection datasets. In \cite{gaus2019evaluation}, two CNNs are used to perform object detection and binary classification (benign or anomaly) in a sequential fashion, where the second CNN takes the localized object within the image as input.
\par
From a safety standpoint, detecting outliers or OOD samples is extremely important and beneficial as training data cannot realistically be large enough to capture all situations. Research in this area is heavily entwined with progress in uncertainty estimation (\cf Chapter \ref{sec:uncertainty}) and domain adaptation (\cf \refsec{subsec:dataset_optimization:domains}). Extending research works to segmentation and object detection tasks would be particularly significant for leveraging autonomous driving research. In addition to safety, OOD detection can be beneficial in other aspects like when using local expert models. For example, when using an expert model for segmentation of urban driving scenes and another expert model for segmentation of highway driving scenes, an OOD detector could act as trigger on which models can be switched. 
\par
With respect to the approaches presented above, uncertainty-based and generative model-based OOD detection methods are currently promising directions of research. However, it remains an open question whether they can unfold their potential well on segmentation and object detection tasks.

\subsection{Active Learning}\label{subsec:dataset_optimization:active_learning}
\sectionauthor{Matthias Rottmann\textsuperscript{5}}
It is widely known that, as a rule of thumb, for the training of any kind of artificial neural network, an increase of training data leads to increased performance.
Obtaining labeled training data, however, is often very costly and time consuming.
\emph{Active learning} provides one possible remedy to this problem: Instead of labeling every data point, active learning utilizes a \emph{query strategy} to request labels from a teacher/an oracle which leverage the model performance most.
The survey paper by Settles \cite{Settles2010} provides a broad overview regarding query strategies for active learning methods. However, except for \emph{uncertainty sampling} and \emph{query by committee}, most of them seem to be infeasible in deep learning applications up to now. Hence, most of the research activities in active deep learning focus on these two query strategies, as we outline in the following.
\par
It has been shown \cite{Gal2017,Matthias2019} for image classification that labels corresponding to uncertain samples can leverage the networks' performance significantly and that a combination with semi-supervised learning is promising. In both works, uncertainty of unlabeled samples is estimated via Monte Carlo (MC) dropout inference. MC dropout inference and a chosen number of training epochs are executed alternatingly, after performing MC dropout inference, the unlabeled samples' uncertainties are assessed by means of sample-wise dispersion measures. Samples for which the DNN model is very uncertain about its prediction are presented to an oracle and labeled.
\par
With respect to object detection, a moderate number of active learning methods has been introduced \cite{Brust2019,Kao2019,Roy2019,Desai2019}.
These approaches include uncertainty sampling \cite{Brust2019,Kao2019} and query-by-committee methods \cite{Roy2019}. In \cite{Kao2019,Desai2019}, additional algorithmic features specifically tailored for object detection networks are presented, \ie separate treatment of the localization and classification loss \cite{Kao2019}, as well as weak and strong supervision schemes \cite{Desai2019}. For semantic segmentation, an uncertainty-sampling-based approach has been presented \cite{Mackowiak2019}, which queries polygone masks for image sections of a fixed size ($128 \times 128$). Queries are performed by means of accumulated entropy in combination with a cost estimation for each candidate image section.
\par
Recently, new methods for estimating the quality of a prediction \cite{devries2018,MetaSeg} as well as new uncertainty quantification approaches, \eg gradient-based ones \cite{Oberdiek2018}, have been proposed. It remains an open question whether they are suitable for active learning. Since most of the conducted studies are rather of academic nature, also their applicability to real-life data acquisition is not yet demonstrated sufficiently. In particular, it is not clear whether the proposed active learning schemes, including the label acquistion, for instance in semantic segmentation, is suitable to be performed by human labelers. Therefore, labeling acquisition with a common understanding of the labelers' convenience and suitability for active learning are a promising direction for research and development.

\subsection{Domains}\label{subsec:dataset_optimization:domains}
\sectionauthor{Julia Rosenzweig\textsuperscript{1}}
The classical assumption in machine learning is that the training and testing data sets are drawn from the same distribution, implying that the model is deployed under the same conditions as it was trained under.  However, as \cite{Moreno-Torres2012,Joshi2012} mention, in real-world applications this assumption is often violated in the sense that the training and the testing set stem from different domains having different distributions. This poses difficulties for statistical models and the performance will mostly degrade when they are deployed on a domain $D^{\mathrm{test}}$, having a different distribution than the training dataset (\ie generalizing from the training to the testing domain is not possible). This makes the study of domains not only relevant from the machine learning perspective, but also from a safety point of view.
\par
More formally, there are differing notions of a 'domain' in literature. For \cite{Csurka2017,Mei2018}, a domain $\mathcal{D} = \{\mathcal{X}, P(X) \}$  consists of a feature space $\mathcal{X} \subset \mathbb{R}^d$ together with a marginal probability distribution $P(X)$ with $X\in \mathcal{X}$. In \cite{Blitzer2007,Ben-David2010}, a domain is a pair consisting of a distribution over the inputs together with a labeling function.
However, instead of a sharp labeling function, it is also widely accepted to define a (training) domain $\mathcal{D} = \{(x_i,y_i)\}_{i=1}^n$ to consist of $n$ (labeled) samples that are sampled from a joint distribution $P(x,y)$ (\cf \cite{Long2018DA}). 
\par
The reasons for \emph{distributional shift} are diverse---as are the names to indicate a shift. For example, if the rate of (class) images of interest is  different between training and testing set this can lead to a domain gap and, \eg result in differing overall error rates. Moreover, as \cite{Chen_2018_CVPR} mentions, changing weather conditions and camera setups in cars lead to a domain mismatch in applications of autonomous driving.
In biomedical image analysis, different imaging protocols and diverse anatomical structures can hinder generalization of trained models (\cf \cite{Dou2019,biomedicalDA17}).
Common terms to indicate distributional shift are \emph{domain shift, dataset shift, covariate shift, concept drift, domain divergence, data fracture, changing environments} or \emph{dataset bias}. References \cite{Storkey2009,Moreno-Torres2012} provide an overview. 
\par
Methods and measures to overcome the problem of domain mismatch between one or more (\cf \cite{MultisourceDA}) source domains and target domain(s) and the resulting poor model performance are studied in the field of transfer learning and in particular its subtopic domain adaptation (\cf \cite{Mei2018}).
For instance, adapting a model that is trained on synthetically generated data to work on real data is one of the core challenges, as can be seen \cite{Chen_2018_CVPR,TakingacloserLook19,VuADVENT2019}.
Furthermore, detecting when samples are out-of-domain or out-of-distribution is an active field of research (\cf \cite{lee2018simple} and the outlier/anomaly detection in \refsec{subsec:dataset_optimization:anomaly_detection} as well as the topic of observers in the black-box methods in \refsec{subsec:verification:black_box_methods} for further reference). This is particularly relevant for machine learning models that operate in the real world: If, \eg an autonomous vehicle encounters some situation that deviates strongly from what was seen during training (\eg due to some special event like a biking competition, carnival, etc.) this can lead to wrong predictions and thereby potential safety issues if not detected in time.

\subsection{Augmentation}\label{subsec:dataset_optimization:augmentation}
\sectionauthor{Falk Kappel\textsuperscript{13}}
Given the need for big amounts of data to train neural networks, one often runs into a situation where data is lacking. This can lead to insufficient generalization and an overfitting to the training data. An overview over different techniques to tackle this challenge can be found in \cite{kukacka2017}. One approach to try and overcome this issue is the augmentation of data. It aims at optimizing available data and increasing its amount, curating a dataset that represents a wide variety of possible inputs during deployment. Augmentation can as well be of help when having to work with a heavily unbalanced dataset by creating more samples of underrepresented classes. A broad survey on data augmentation is provided by \cite{shorten2019}. They distinguish between two general approaches to data augmentation with the first one being data warping augmentations that focus on taking existing data and transforming it in a way that does not effect labels. The other option are oversampling augmentations, which create synthetic data that can be used to increase the size of the dataset.\newline
Examples of some of the most basic augmentations are flipping, cropping, rotating, translating, shearing and zooming. These are affecting the geometric properties of the image and are easily implemented \cite{shorten2019}. The machine learning toolkit \texttt{Keras}, for example, provides an easy way of applying them to data using their \texttt{ImageDataGenerator} class \cite{chollet2015keras}. Other simple methods include adaptations in color space that affect properties such as lighting, contrast and tints, which are common variations within image data. Filters can be used to control increased blur or sharpness \cite{shorten2019}. In \cite{Zhong2017} random erasing is introduced as a method with similar effect as cropping, aiming at gaining robustness against occlusions. An example for mixing images together as an augmentation technique can be found in \cite{inoue2018}. \newline
The abovementioned methods have in common that they work on the input data but there are different approaches that make use of deep learning for augmentation. An example for making augmentations in feature space using autoencoders can be found in \cite{devries2017}. They use the representation generated by the encoder and generate new samples by interpolation and extrapolation between existing samples of a class. The lack of interpretability of augmentations in feature space in combination with the tendency to perform worse than augmentations in image space present open challenges for those types of augmentations \cite{shorten2019, wong2016}. Adversarial training is another method that can be used for augmentation. The goal of adversarial training is to discover cases that would lead to wrong predictions. That means the augmented images won't necessarily represent samples that could occur during deployment but that can help in achieving more robust decision boundaries \cite{shorten2019}. An example of such an approach can be found in \cite{chen2018b}. Generative modelling can be used to generate synthetic samples that enlarge the dataset in a useful way with GANs, variational autoencoders and the combination of both are important tools in this area \cite{shorten2019}. Examples for data augmentation in medical context using a CycleGAN \cite{zhu2017} can be found in \cite{sandfort2019} and using a progressively growing GAN \cite{karras2017} in \cite{bowles2018}. Next to neural style transfer \cite{gatys2015} that can be used to change the style of an image to a target style, AutoAugment \cite{Cubuk2019} and population based augmentation \cite{ho2019} are two more interesting publications. In the latter two, the idea is to search a predefined search space of augmentations to gather the best selection.
\subsection{Corner Case Detection}\label{subsec:dataset_optimization:corner_case_detection}
\sectionauthor{Alexander Pohl\textsuperscript{16}, Marco Hoffmann\textsuperscript{16}, Michael Mlynarski\textsuperscript{16}, Timo Sämann\textsuperscript{4}}
Ensuring that AI-based applications behave correctly and predictably even in unexpected or rare situations is a major concern that gains importance especially in safety-critical applications such as autonomous driving. In the pursuit of more robust AI corner cases play an important role. 
\par
The meaning of the term corner case varies in the literature. Some consider mere erroneous or incorrect behavior as corner cases \cite{zhang:2019, tian:2018, pei:2017}. For example, in \cite{bolte:2019} corner cases are referred to as situations in which an object detector fails to detect relevant objects at relevant locations. Others characterize corner cases mainly as rare combinations of input parameter values \cite{hanhirova:2020, koopman:2018}. This project adopts the first definition: Inputs that result in unexpected or incorrect behaviour of the AI function are defined as corner cases. 
\par
Contingent on the hardware, the AI architecture and the training data, the search space of corner cases quickly becomes incomprehensibly large. While manual creation of corner cases (\eg constructing or re-enacting scenarios) might be more controllable, approaches that scale better and allow for a broader and more systematic search for corner cases require extensive automation.
\par
One approach to automatic corner case detection is based on transforming the input data. The \emph{DeepTest} framework \cite{tian:2018} uses three types of image transformations: linear, affine and convolutional transformations. In addition to these transformations, metamorphic relations help detect undesirable behaviors of deep learning systems. They allow changing the input while asserting some characteristics of the result \cite{xie:2011}. For example, changing the contrast of input frames should not affect the steering angle of a car \cite{tian:2018}. Input-output pairs that violate those metamorphic relations can be considered as corner cases.
\par
Among other things, the white-box testing framework \emph{DeepXplore} \cite{pei:2017} applies a method called \emph{gradient ascent} to find corner cases (\cf \refsec{subsec:verification:formal_testing}). In the experimental evaluation of the framework, three variants of deep learning architectures were used to classify the same input image. The input image was then changed according to the gradient ascent of an objective function that reflected the difference in the resulting class probabilities of the three model variants. When the changed (now artificial) input resulted in different class label predictions by the model variants, the input was considered as a corner case.  
\par
In \cite{bolte:2019}, corner cases are detected on video sequences by comparing predicted with actual frames. The detector has three components: The first component, semantic segmentation, is used to detect and locate objects in the input frame. As the second component, an image predictor trained on frame sequences predicts the actual frame based on the sequence preceding that  frame. An error is determined by comparing the actual with the predicted (\ie expected) frame, following the idea that only situations that are unexpected for AI-based perception functions may be potentially dangerous and therefore a corner case. Both the segmentation and the prediction error are then fed into the third component of the detector, which determines a corner case score that reflects the extent to which unexpected relevant objects are at relevant locations.
\par
In \cite{hanhirova:2020}, a corner case detector based on simulations in a \emph{Carla} environment \cite{dosovitskiy2017carla} is presented. In the simulated world, AI agents control the vehicles. During simulations, state information of both the environment and the AI agents are fed into the corner case detector. While the environment provides the real vehicle states, the AI agents provide estimated and perceived state information. Both sources are then compared to detect conflicts (\eg collisions). These conflicts are recorded for analysis. 
\par
Several ways of automatically generating and detecting corner cases exist. However, corner case detection is a task with challenges of its own: Depending on the operational domain including its boundaries, the space of possible inputs can be very large. Also, some types of corner cases are specific to the AI architecture, \eg the network type or the network layout used. Thus, corner case detection has to assume a holistic point of view on both model and input, adding further complexity and reducing transferability of previous insights. 
\par
Although it can be argued that rarity does not necessarily characterize corner cases, rare input data might have the potential of challenging the AI functionality (\cf \refsec{subsec:dataset_optimization:anomaly_detection}).
Another research direction could investigate whether structuring the input space in a way suitable for the AI functionality supports the detection of corner cases. Provided that the operational domain is conceptualized as an ontology, ontology-based testing \cite{bagschik:2018} may support automatic detection. A properly adapted generator may specifically select promising combinations of extreme parameter values and, thus, provide valuable input for synthetic test data generation. 
\section{Robust Training}\label{sec:robust_training}
\sectionauthor{Nikhil Kapoor\textsuperscript{7}}
Recent works~\cite{Azulay2018, Hendrycks2019, Rodner2016, Engstrom2019, Bunne2018, Baer2019, Fawzi2015} have shown that state-of-the-art deep neural networks (DNNs) performing a wide variety of computer vision tasks such as image classification~\cite{Krizhevsky2012, He2015, Mahajan2018}, object detection~\cite{Girshick2015, Redmon2015, He2017} and semantic segmentation~\cite{Chen2017, Zhu2019, Wang2019, Loehdefink2019} are not robust to small changes in the input.
\par
Robustness of neural networks is an active and open research field that can be considered highly relevant for achieving safety in autonomous driving. Currently, most of the research is directed towards either improving adversarial robustness~\cite{Szegedy2014} (robustness against carefully designed perturbations that aim at causing misclassifications with high confidence), or improving corruption robustness~\cite{Hendrycks2019} (robustness against commonly occurring augmentations such as weather changes, addition of Gaussian noise, photometric changes, etc.). While adversarial robustness might be more of a security issue than a safety issue, corruption robustness, on the other hand, can be considered highly safety-relevant.
\par
Equipped with these definitions, we broadly term \emph{robust training} here as methods or mechanisms that aim at improving either adversarial or corruption robustness of a DNN, by incorporating modifications into the architecture or into the training mechanism itself.
\subsection{Hyperparameter Optimization} \label{subsec:robust_training:hyperparameter_optimization}
\sectionauthor{Seyed Eghbal Ghobadi\textsuperscript{8}, Patrick Feifel\textsuperscript{8}}
The final performance of a neural network depends highly on the learning process. The process includes the actual optimization and may additionally introduce training methods such as dropout, regularization, or parametrization of a multi-task loss. 
\par
These methods adapt their behavior for predefined parameters. Hence, their optimal configuration is a priori unknown. We refer to them as \emph{hyperparameters}. Important hyperparameters comprise, for instance, the initial learning rate, steps for learning rate reduction, learning rate decay, momentum, batch size, dropout rate and number of iterations. Their configuration has to be determined according to the architecture and task of the CNN \cite{hutter_automated_2019}. The search of an optimal hyperparameter configuration is called hyperparameter optimization (HO).
\par
HO is usually described as an optimization problem \cite{hutter_automated_2019}. Thereby, the combined configuration space is defined as 
$\boldsymbol{\Lambda} = \lambda_1 \times \lambda_2 \times \cdots \lambda_N$,
according to each domain $\lambda_n$. Their individual spaces can be continuous, discrete, categorical or binary. 
\par
Hence, we aim to find an optimal hyperparameter configuration $\lambda^{\star}$ by minimizing an objective function $\mathcal{O}\left ( \right )$, which evaluates a trained model $\mathcal{M}$ on the validation dataset $\mathcal{D}^{\textrm{val}}$ with the loss $\mathcal{L}$:
\begin{equation}
\lambda^{\star} = 
\text{arg  min}_{\lambda \in \boldsymbol{\Lambda}} \; 
\mathcal{O} \left( \mathcal{L}, \mathcal{M}_{\lambda}, \mathcal{D}^{\textrm{train}}, \mathcal{D}^{\textrm{val}} \right)
\end{equation}
This problem statement is widely regarded in traditional machine learning and primarily based on Bayesian optimization (BO) in combination with Gaussian processes. However, a straightforward application to deep neural networks encounters problems due to a \emph{lack of scalability, flexibility and robustness} \cite{zhang_deep_2019}, \cite{falkner_bohb_2018}. 
\par
To exploit the benefits of BO, many authors proposed different combinations with other approaches. 
Hyperband \cite{li_hyperband_2017} in combination with BO (BOHB) \cite{falkner_bohb_2018} frames the optimization as ``... a pure exploration non-stochastic infinite-armed bandit problem ...''. 
The method of BO for iterative learning (BOIL) \cite{nguyen_bayesian_2019} internalizes iteratively collected information about the learning curve and the learning algorithm itself. 
The authors of \cite{wu_practical_2019} introduce the trace-aware knowledge gradient (taKG) as an acquisition function for BO (BO-taKG) which ``leverages both trace information and multiple fidelity controls''.
Thereby BOIL and BO-taKG achieve state-of-research performance regarding CNNs outperforming Hyperband.
\par
Other approaches such as the orthogonal array tuning method (OATM) \cite{zhang_deep_2019} or HO by reinforcement learning (Hyp-RL) \cite{jomaa_hyp-rl_2019} turn away from the Bayesian approaches and offer new research directions.
\par
Finally, the insight that many authors include kernel sizes and number of kernels and layers in their hyperparameter configuration should be emphasized. More work should be spent on the distinct integration of HO in the performance estimation strategy of neural architecture search (\cf \refsec{subsec:architecture:neural_architecture_search}).

\subsection{Modification of Loss} \label{subsec:robust_training:modification_of_loss}
\sectionauthor{Nikhil Kapoor\textsuperscript{7}}
There exist many approaches that aim at directly modifying the loss function with an objective of improving either adversarial or corruption robustness~\cite{Saito2020, Li2019, Xu2019, Seck2019, Pang2018, Kumar2018, Wang2019}. One of the earliest approaches for improving corruption robustness was introduced by Zheng \etal \cite{zheng2016} called \emph{stability training}, where they introduce a regularization term that penalizes the network prediction to a clean and an augmented image. However, their approach does not scale to many augmentations at the same time. Janocha \etal \cite{Janocha2017} then introduced a detailed analysis on the influence of multiple loss functions to model performance as well as robustness and suggested that expectation-based losses tend to work better with noisy data and squared-hinge losses tend to work better for clean data. Other well-known approaches are mainly based on variations of data augmentation~\cite{Cubuk2019, Cubuk2019_1, Cubuk2019_2, Cubuk2019_3}, which can be computationally quite expensive. 
\par
In contrast to corruption robustness, there exist many more approaches based on adversarial examples. We highlight some of the most interesting and relevant ones here. Mustafa \etal \cite{Mustafa2019} proposes to add a loss term that maximally separates class-wise feature map representations, hence increasing the distance from data points to the corresponding decision boundaries. Similarly, Pang \etal \cite{Pang2020} proposed the Max-Mahalanobis center (MMC) loss to learn more structured representations and induce high-density regions in the feature space. Chen \etal \cite{Chen2018} proposed a variation of the well-known cross entropy (CE) loss that not only maximizes the model probabilities of the correct class, but in addition, also minimizes model probabilities of incorrect classes. Cisse \etal \cite{Cisse2017} constraints the Lipschitz constant of different layers to be less than one which restricts the error propagation introduced by adversarial perturbations to a DNN. Dezfooli \etal \cite{Moosavi-Dezfooli2019} proposed to minimize the curvature of the loss surface locally around data points. They emphasize that there exists a strong correlation between locally small curvature and correspondingly high adversarial robustness. 
\par
All of these methods highlighted above are evaluated mostly for image classification tasks on smaller datasets, namely CIFAR-10~\cite{Krizhevsky2009}, CIFAR-100~\cite{Krizhevsky2009}, SVHN~\cite{Netzer2011}, and only sometimes on ImageNet~\cite{Krizhevsky2012}. Very few approaches have been tested rigorously on complex safety-relevant tasks such as \emph{object detection} and \emph{semantic segmentation}, etc. Moreover, methods that improve adversarial robustness are only tested on a small subset of attack types under differing attack specifications. This makes comparing multiple methods difficult. 
\par
In addition, methods that improve corruption robustness are evaluated over a standard data set of various corruption types which may or may not be relevant to its application domain. In order to assess multiple methods for their effect on safety-related aspects, a thorough robustness evaluation methodology is needed, which is largely missing in the current literature. This evaluation would need to take into account relevant disturbances/corruption types present in the real world (application domain) and had to assess robustness towards such changes in a rigorous manner. Without such an evaluation, we run the risk of being overconfident in our network, thereby harming safety.
\subsection{Domain Generalization}
\sectionauthor{Firas Mualla\textsuperscript{13}}
Domain generalization (DG) can be seen as an extreme case of \emph{domain adaptation} (DA). The latter is a type of transfer learning, where the source and target tasks are the same (\eg shared class labels) but the source and target domains are different (\eg another image acquisition protocol or a different background) \cite{Csurka17:SurveyDA,Wang20:SurveyFSL}. 
The DA can be either supervised (SDA), where there is little available labeled data in the target domain, or unsupervised (UDA), where data in the target domain is not labeled. The DG goes one step further by assuming that the target domain is entirely unknown. Thus, it seeks to solve the train-test domain shift in general. While DA is already an established line of research in the machine learning community, DG is relatively new \cite{MuandetBS13:FirstDG}, though with an extensive list of papers in the last few years. 
\par
Probably, the first intuitive solution that one may think of to implement DG is neutralizing the domain-specific features. It was shown in \cite{Wang19:glcm} that the gray-level co-occurrence matrices (GLCM) tend to perform poorly in semantic classification (\eg digit recognition) but yield good accuracy in textural classification compared to other feature sets such as SURF and LBP. DG was thus implemented by decorrelating the model's decision from the GLCM features of the input image even without the need of domain labels.

Besides the aforementioned intensity-based statistics of an input image, it is known that characterizing image style can be done based on the correlations between the filter responses of a DNN layer \cite{Gatys16:NST} (neural style transfer). In \cite{Somavarapu20:StylizedDG}, the training images are enriched with stylized versions, where a style is defined either by an external style (\eg cartoon or art) or by an image from another domain. Here, DG is addressed as a \emph{data augmentation} problem.

Some approaches \cite{Li18:AdversarialDG, Matsuura20:NoDomainLabel} try to learn generalizable latent representations by a kind of adversarial training. This is done by a generator or an encoder, which is trained to generate a hidden feature space that maximizes the error of a domain discriminator but at the same time minimizes the classification error of the task of concern. Another flavor of adversarial training can be seen in \cite{Li18:AAE-MMD-DG}, where an adversarial autoencoder \cite{Makhzani16:AAE} is trained to generate features, which a discriminator cannot distinguish from random samples drawn from a prior Laplace distribution. This regularization prevents the hidden space from overfitting to the source domains, in a similar spirit to how variational autoencoders do not leave gaps in the latent space. In \cite{Matsuura20:NoDomainLabel}, it is argued that the domain labels needed in such approaches are not always well-defined or easily available. Therefore they assume unknown latent domains which are learned by clustering in a space similar to the style-transfer features mentioned above. The pseudo labels resulting from clustering are then used in the adversarial training.

Autoencoders have been employed for DG not only in an adversarial setup, but also in the sense of \emph{multi-task learning} nets \cite{Caruana1997}, where the classification task in such nets is replaced by a reconstruction one. In \cite{Ghifary15:MTAE}, an autoencoder is trained to reconstruct not only the input image but also the corresponding images in the other domains.

In the core of both DA and DG we are confronted with a distribution matching problem. However, estimating the probability density in high-dimensional spaces is intractable. The density-based metrics such as Kullback-Leibler divergence are thus not directly applicable. In statistics, the so-called \emph{two-samples tests} are usually employed to measure the distance between two distributions in a point-wise manner, \ie without density estimation. For deep learning applications, these metrics need not only to be point-wise but also differentiable. The two-samples tests were approached in the machine learning literature using (differentiable) K-NNs \cite{Djolonga17:Differentiable-Knn}, classifier two-samples tests (C2ST) \cite{Lopez-PazO17:C2ST}, or based on the theory of kernel methods \cite{Smola07:KernelForDistributions}. More specifically, the \emph{maximum mean discrepancy} (MMD) \cite{Gretton06:MMD,Gretton12:MMD-Journal}, which belongs to the kernel methods, is widely used for DA \cite{Ghifary14:MMD-DA, Long17:MMD-DA, Yan17:MMD-UDA, Yan20:MMD-UDA} but also for DG \cite{Li18:AAE-MMD-DG}. Using the MMD, the distance between two samples is estimated based on pairwise kernel evaluations, \eg the radial basis function (RBF) kernel.

While the DG approaches generalize to domains from which zero shots are available, the so-called \emph{zero shot learning} (ZSL) approaches generalize to tasks (\eg new classes in the same source domains) for which zero shots are available. Typically, the input in ZSL is mapped to a semantic vector per class instead of a simple class label. This can be, for instance, a vector of visual attributes \cite{Lampert14:ABZSL} or a word embedding of the class name \cite{Kodirov17:AEZSL}. A task (with zero shots at training time) can be then described by a vector in this space. In \cite{Mancini20:ZSL-DG}, there is an attempt to combine ZSL and DG in the same framework in order to generalize to new domains as well as new tasks, which is also referred to as
 \emph{heterogeneous domain generalization}.

Note that most discussed approaches for DG require non-standard handling, \ie modifications to models, data, and/or the optimization procedure. This issue poses a serious challenge as it limits the practical applicability of these approaches. There is a line of research which tries to address this point by linking DG to other machine learning paradigms, especially the model-agnostic meta-learning (MAML) \cite{FinnAL17:MAML} algorithm, in an attempt to apply DG in a model-agnostic way. Loosely speaking, a model can be exposed to simulated train-test domain shift by training on a small \emph{support set} to minimize the classification error on a small \emph{validation set}. This can be seen as an instance of a \emph{few shot learning} (FSL) problem \cite{Wang20:SurveyFSL}. Moreover, the procedure can be repeated on other (but related) FSL tasks (\eg different classes) in what is known as \emph{episodic training}. The model transfers its knowledge from one task to another task and learns how to learn fast for new tasks. This can be thus seen as a \emph{meta-learning} objective \cite{Hospedales20:SurveyML} (in a FSL setup). Since the goal of DG is to adapt to new domains rather than new tasks, several model-agnostic approaches \cite{Li18:ML-DG, Li19:EpisodicDG,Balaji18:ML-DG,Dou19:ModelAgnosticDG} try to recast this procedure in a DG setup.
\section{Adversarial Attacks}\label{sec:adversarial_attacks}
\sectionauthor{Andreas Bär\textsuperscript{15}}
Over the last few years, deep neural networks (DNNs) consistently showed state-of-the-art performance across several vision-related tasks.
While their superior performance on clean data is indisputable, they show a lack of robustness to certain input patterns, denoted as \emph{adversarial examples} \cite{Szegedy2014}.
In general, an algorithm for creating adversarial examples is referred to as an \emph{adversarial attack} and aims at fooling an underlying DNN, such that the output changes in a desired and malicious way.
This can be carried out without any knowledge about the DNN to be attacked (black-box attack) \cite{Moosavi-Dezfooli2016, Papernot2017}, or with full knowledge about the parameters, architecture, or even training data of the respective DNN (white-box attack) \cite{Goodfellow2015, Carlini2017, Madry2018}.
While initially being applied on simple classification tasks, some approaches aim at finding more realistic attacks \cite{thys2019fooling, jia2019fooling}, which particularly pose a threat to safety-critical applications, such as DNN-based environment perception systems in autonomous vehicles.
Altogether, this motivated the research in finding ways of defending against such adversarial attacks \cite{Goodfellow2015, Moosavi-Dezfooli2019, Xie2019a, Guo2018}.
In this section, we introduce the current state of research regarding adversarial attacks in general, more realistic adversarial attacks closely related to the task of environment perception for autonomous driving, and strategies for detecting or defending adversarial attacks.
We conclude each section by clarifying current challenges and research directions.  
\subsection{Adversarial Attacks and Defenses}\label{subsec:adverarial_attacks:adversarial_attacks_defenses}
\sectionauthor{Andreas Bär\textsuperscript{15}, Seyed Eghbal Ghobadi\textsuperscript{8}, Ahmed Hammam\textsuperscript{8}}
The term \emph{adversarial example} was first introduced by Szegedy \etal \cite{Szegedy2014}.
From there on, many researchers tried to find new ways of crafting adversarial examples more effectively.
Here, the fast gradient sign method (FGSM) \cite{Goodfellow2015}, DeepFool \cite{Moosavi-Dezfooli2016}, least-likely class method (LLCM) \cite{Kurakin2017,Kurakin2017a}, C\&W \cite{Carlini2017a}, momentum iterative fast gradient sign method (MI-FGSM) \cite{Dong2018a}, and projected gradient descent (PGD) \cite{Madry2018} are a few of the most famous attacks so far.
In general, these attacks can be executed in an iterative fashion, where the underlying adversarial perturbation is usually bounded by some norm and is following additional optimization criteria, \eg minimizing the number of changed pixels.
\par
The mentioned attacks can be further categorized as image-specific attacks, where for each image a new perturbation needs to be computed.
On the other hand, image-agnostic attacks aim at finding a perturbation, which is able to fool an underlying DNN on a set of images.
Such a perturbation is also referred to as a \emph{universal adversarial perturbation} (UAP).
Here, the respective algorithm UAP \cite{Moosavi-Dezfooli2017}, fast feature fool (FFF) \cite{Mopuri2017}, and prior driven uncertainty approximation (PD-UA) \cite{Liu2019e} are a few honorable mentions.
Although the creation process of a universal adversarial perturbation typically relies on a white-box setting, they show a high \emph{transferability} across models \cite{Hashemi2020}.
This allows black-box attacks, where one model is used to create a universal adversarial perturbation, and another model is being attacked with the beforehand-created perturbation.
Another way of designing black-box attacks is to create a surrogate DNN, which mimics the respective DNN to be attacked and thus can be used in the process of adversarial example creation \cite{Papernot2017}.
On the contrary, some research has been done to create completely incoherent images (based on evolutionary algorithms or gradient ascent) to fool an underlying DNN \cite{Nguyen2015}.
Different from that, another line of work has been proposed to alter only some pixels in images to attack a respective model.
Here \cite{narodytska2016simple} and \cite{narodytska2017simple} have used optimization approaches to perturb some pixels in images to produce targeted attacks, aiming at a specific class output, or non-targeted attacks, aiming at outputting a class different from the network output or the ground truth.
This can be extended up to finding one pixel in the image to be exclusively perturbed to generate adversarial images \cite{su2019one,narodytska2016simple}.
The authors of \cite{baluja2017adversarial,poursaeed2018generative,sarkar2017upset} proposed to train generative models to generate adversarial examples.
Given an input image and the target label, a generative model is trained to produce adversarial examples for DNNs.
However, while the produced adversarial examples look rather unrealistic to a human, they are able to completely deceive a DNN.
\par
The existence of adversarial examples not only motivated research in finding new attacks, but also in finding \emph{defense strategies} to effectively defend these attacks.
Especially for safety-critical applications, such as DNN-based environment perception for autonomous driving, the existence of adversarial examples needs to be handled accordingly.
Similar to adversarial attacks, one can categorize defense strategies into two types: \emph{model-specific} defense strategies and \emph{model-agnostic} defense strategies.
The former refers to defense strategies, where the model of interest is modified in certain ways.
The modification can be done on the architecture, training procedure, training data, or model weights.
On the other hand, model-agnostic defense strategies consider the model to be a black box.
Here, only the input or the output is accessible.
Some well-known model-specific defense strategies include adversarial training \cite{Goodfellow2015, Madry2018}, the inclusion of robustness-oriented loss functions during training \cite{Chen2019e, Moosavi-Dezfooli2019, Kapoor2020}, removing adversarial patterns in features by denoising layers \cite{He2019c,Mustafa2019a, Xie2019}, and redundant teacher-student frameworks \cite{Baer2019, Baer2020}.
The majority of model-agnostic defense strategies primarily focuses on various kinds of (gradient masking) pre-processing strategies \cite{Bai2019, Guo2018,Gupta2019, Jia2019,Liu2019c,Raff2019,Theagarajan2019}.
The idea is to remove the adversary from the respective image, such that the image is transformed from the adversarial space back into the clean space.
\par
Nonetheless, Athalye \etal \cite{Athalye2018} showed that gradient masking alone is not a sufficient criterion for a reliable defense strategy.
In addition, detection and out-of-distribution techniques have also been proposed as model-agnostic defense strategies against adversarial attacks.
Here, the Mahalanobis distance \cite{lee2018simple} or area under the receiver operating characteristic curve (AUROC) and area under the precision-recall curve (AUPR) \cite{hendrycks2017baseline} are used to detect adversarial examples.
The authors of \cite{hendrycks2017baseline,lee2017training,Metzen2017} on the other hand proposed to train networks to detect, whether the input image is out-of-distribution or not. 
\par
Moreover, Feinman \etal\cite{feinman2017detecting} proved that adversarial attacks usually produce high uncertainty on the output of the DNN.
As a consequence, they proposed to use the dropout technique to estimate uncertainty on the output to identify a possible adversarial attack.
\par
Regarding adversarial attacks, the majority of the listed attacks are designed for image classification.
Only a few adversarial attacks consider tasks that are closely related to autonomous driving, such as bounding box detection, semantic segmentation, instance segmentation, or even panoptic segmentation.
Also, the majority of the adversarial attacks rely on a white-box setting, which is usually not present for a potential attacker.
Especially universal adversarial perturbations have to be considered as a real threat due to their high model transferability.
Generally speaking, the existence of adversarial examples has not been thoroughly studied yet.
An analytical interpretation is still missing, but could help in designing more mature defense strategies.
\par
Regarding defense strategies, adversarial training is still considered as one of the most effective ways of increasing the robustness of a DNN.
Nonetheless, while adversarial training is indeed effective, it is rather inefficient in terms of training time.
In addition, model-agnostic defenses should be favored as once being designed, they can be easily transferred to different models.
Moreover, as most model-agnostic defense strategies rely on gradient-masking and it has been shown that gradient-masking is not a sufficient property for a defense strategy, new ways of designing model-agnostic defenses should be taken into account.
Furthermore, out-of-distribution and adversarial attacks detection or even correction methods have been a new trend for identifying attacks.
However, as the environment perception system of an autonomous driving vehicle could rely on various information sources, including LiDAR, optical flow, or depth from a stereo camera, techniques of information fusion should be further investigated to mitigate or even eliminate the effect of adversarial examples.

\subsection{More Realistic Attacks}\label{subsec:adverarial_attacks:more_realistic_attacks}
\sectionauthor{Svetlana Pavlitskaya\textsuperscript{14}}

We consider the following two categories of realistic adversarial attacks: (1) image-level attacks, which not only fool a neural network but also pose a provable threat to autonomous vehicles, and (2) attacks which have been applied in a real world or in a simulation environment, such as car learning to act (CARLA) \cite{dosovitskiy2017carla}.
\par
Some notable examples in the first category of attacks include attacks on semantic segmentation \cite{Metzen2017} or person detection \cite{thys2019fooling}.
\par
In the second group of approaches, the attacks are specifically designed to survive real world distortions, including different distances, weather and lighting conditions, as well as camera angles. For this, adversarial perturbations are usually concentrated in a specific image area, called \emph{adversarial patch}. Crafting an adversarial patch involves specifying a patch region in each training image, applying transformations to the patch, and iteratively changing the pixel values within this region to maximize the network prediction error. The latter step typically relies on an algorithm, proposed for standard adversarial attacks, which aim at crafting invisible perturbations while misleading neural networks, \eg C\&W \cite{Carlini2017a}, Jacobian-based saliency map attack (JSMA) \cite{Papernot2017}, and PGD \cite{Madry2018}.
\par
The first printable adversarial patch for image classification was described by Brown \etal \cite{brown2017adversarial}. Expectation over transformations (EOT) \cite{athalye2017synthesizing} is one of the influential updates to the original algorithm---it permits to robustify patch-based attacks to distortions and affine transformations. Localized and visible adversarial noise (LaVAN) \cite{karmon2018lavan} is a further method to generate much smaller patches (up to 2\% of the pixels in the image). In general, fooling image classification with a patch is a comparatively simple task, because adversarial noise can mimic an instance of another class and thus lower the prediction probability for a true class.
\par
Recently, patch-based attacks for a more tricky task of object detection have been described \cite{liu2018dpatch,thys2019fooling}. Also, Lee and Kolter \cite{lee2019physical} generate a patch using PGD \cite{Madry2018}, followed by EOT applied to the patch. With this approach, all detections in an image can be successfully suppressed, even without any overlap of a patch with bounding boxes. Furthermore, several approaches for generating an adversarial T-shirt have been proposed, including \cite{xu2019adversarial,wu2019making}.
\par
DeepBillboard \cite{zhou2018deepbillboard} is the first attempt to attack end-to-end driving models with adversarial patches. The authors propose to generate a single patch for a sequence of input images to mislead four steering models, including DAVE-2 in a drive-by scenario. 
\par
Apart from physical feasibility, inconspicuousness is crucial for a realistic attack. Whereas adversarial patches usually look like regions of noise, several works have explored attacks with an inconspicuous patch. In particular, Eykholt \etal\cite{eykholt2018robust} demonstrate the vulnerability of road sign classification to the adversarial perturbations in the form of only black and white stickers. In \cite{boloor2019simple}, an end-to-end driving model is attacked in CARLA by  painting  of  black lines on the road. Also, Kong and Liu \cite{kong2019generating} use a generative adversarial network to get a realistic billboard to attack an end-to-end driving model in a drive-by scenario. In \cite{duan2020adversarial}, a method to hide visible adversarial perturbations with customized styles is proposed, which leads to adversarial traffic signs that look unsuspicious to a human.
\par
Current research mostly focuses on attacking image-based perception of an autonomous vehicle. Adversarial vulnerability of further components of an autonomous vehicle, \eg LiDAR-based perception, optical flow and depth estimation, has only recently gained attention. Furthermore, most attacks consider only a single component of an autonomous driving pipeline, the question whether the existing attacks are able to propagate to further pipeline stages has not been studied yet. The first work in this direction \cite{jia2019fooling} describes an attack on object detection and tracking. The evaluation is, however, limited to a few clips, where no experiments in the real world have been performed. Overall, the research on realistic adversarial attacks, especially combined with physical tests, is currently in the starting phase.
\section{Interpretability}\label{sec:interpretability}
\sectionauthor{Felix Brockherde\textsuperscript{10}}
Neural networks are, by their nature, black boxes and therefore intrinsically hard to interpret \cite{taylor2006methods}. Due to their unrivaled performance, they still remain first choice for advanced systems even in many safety-critical areas, such as level 4 automated driving. This is why the research community has invested considerable effort to unhinge the black-box character and make deep neural networks more transparent.

We can observe three strategies that provide different view points towards this goal in the state of the art. First is the most direct approach of opening up the black box and looking at intermediate representations (\refsec{subsec:interpretability:intermediate_representations}). Being able to interpret individual layers of the system facilitates interpretation of the whole. The second approach tries to provide interpretability by explaining the network's decisions with pixel attributions (\refsec{subsec:interpretability:pixel_attribution}). Aggregated explanations of decision can then lead to interpretability of the system itself. Third is the idea of approximating the network with interpretable proxies to benefit from the deep neural networks performance while allowing interpretation via surrogate models (\refsec{subsec:interpretability:interpretable_proxies}). Underlying all aspects here is the area of visual analytics (\refsec{subsec:interpretability:visual_analytics}).

There exists earlier research in the medical domain to help human experts understand and convince them of machine learning decisions \cite{caruana2015intelligible}. Legal requirements in the finance industry gave rise to interpretable systems that can justify their decisions. An additional driver for interpretability research was the concern for Clever Hans predictors \cite{lapuschkin2019unmasking}.

\subsection{Visual Analytics}\label{subsec:interpretability:visual_analytics}
\sectionauthor{Elena Schulz\textsuperscript{1}}
Traditional data science has developed a huge tool set of automated analysis processes conducted by computers, which are applied to problems that are well-defined in the sense that the dimensionality of input and output as well as the size of the data set they rely on is manageable. 
For those problems that in comparison are more complex, the automation of the analysis process is limited and/or might not lead to the desired outcome. 
This is especially the case with unstructured data like image or video data in which the underlying information cannot directly be expressed by numbers. 
Rather, it needs to be transformed to some structured form to enable computers to perform some task of analysis. 
Additionally, with an ever increasing amount of various types of data being collected, this ``information overload'' cannot solely be analyzed by automatic methods \cite{Keim2008, Keim2009}. 
\par
\emph{Visual analytics} addresses this challenge as ``the science of analytical reasoning facilitated by interactive visual interfaces'' \cite{cook_2005}. 
Visual analytics therefore does not only focus on either computationally processing data or visualizing results but coupling both tightly with interactive techniques.  
Thus, it enables an integration of the human expert into the iterative visual analytics process: Through visual understanding and human reasoning, the knowledge of the human expert can be incorporated to effectively refine the analysis. This is of particular importance, where a stringent safety argumentation for complex models is required. With the help of visual analytics, the line of argumentation can be built upon arguments that are understandable for humans.
To include the human analyst efficiently into this process, a possible guideline is the \emph{visual analytics mantra} by Keim: ``Analyze first, show the important, zoom, filter and analyze further, details on demand'' \cite{Keim2008}\footnote{Extending the original \emph{visualization mantra} by Shneiderman ``Overview first, filter and zoom, details on demand''\cite{Shneiderman}.}.
\par
The core concepts of visual analytics therefore rely on well-designed interactive visualizations, which support the analyst in the tasks of, \eg reviewing, understanding, comparing and inferring not only the initial phenomenon or data but also the computational model and its results itself with the goal of enhancing the analytical process. 
\par
Driven by various fields of application, visual analytics is a multidisciplinary field with a wide variety of task-oriented development and research. 
As follows, recent work has been done in several areas: 
depending on the task, there exist different pipeline approaches to create whole \emph{visual analytics systems} \cite{Wang2016}; 
the injection of human expert knowledge into the process of determining trends and patterns from data is the focus of \emph{predictive visual analytics} \cite{Lu2017b,Lu2017a}; 
enabling the human to explore \emph{high-dimensional data} \cite{Liu2017b} interactively and visually (\eg via dimensionality reduction \cite{Sacha2017a}) is a major technique  to enhance the understandability of complex models (\eg neural networks); 
the iterative improvement and the understanding of machine learning models is addressed by using interactive visualizations in the field of \emph{general machine learning} \cite{Liu2017a} or the other way round: using machine learning to improve visualizations and guidance based on user interactions \cite{Endert2017}.
Even more focused on the loop of simultaneously developing and refining machine learning models is the area of \emph{interactive machine learning}, where the topics of \emph{interface design} \cite{Dudley2018} and the \emph{importance of users} \cite{Amershi2014,Sacha2017b} are discussed. One of the current research directions is using visual analytics in the area of \emph{deep learning} \cite{Garcia2018,Hohman2018,Choo2018}. However, due to the interdisciplinarity of visual analytics, there are still open directions and ongoing research opportunities. 
\par
Especially in the domain of neural networks and deep learning, visual analytics is a relatively new approach in tackling the challenge of \emph{explainability} and \emph{interpretability} of those often called \emph{black boxes}. 
To enable the human to better interact with the models, research is done in enhancing the \emph{understandability} of complex deep learning models and their outputs with the use of proper visualizations. 
Other research directions attempt to achieve improving the \emph{trustability} of the models, giving the opportunity to inspect, diagnose and refine the model. 
Further, possible areas for research are \emph{online training processes} and the development of \emph{interactive systems} covering the whole process of training, enhancing and monitoring machine learning models. 
Here, the approach of \emph{mixed guidance}, where system-initiated guidance is combined with user-initiated guidance, is discussed among the visual analytics community as well. 
Another challenge and open question is creating ways of \emph{comparing models} to examine which model yields a better performance, given specific situations and selecting or combining the best models with the goal of increasing performance and overall safety.

\subsection{Intermediate Representations}\label{subsec:interpretability:intermediate_representations}
\sectionauthor{Felix Hauser\textsuperscript{11}, Jan Kronenberger\textsuperscript{9}, Seyed Eghbal Ghobadi\textsuperscript{8}}
In general, representation learning \cite{bengioRepresentationLearningReview2013} aims to extract lower dimensional features in latent space from higher dimensional inputs.
These features are then used as an effective representation for regression, classification, object detection and other machine learning tasks.
Preferably, latent features should be disentangled, meaning that they represent separate factors found in the data that are statistically independent.
Due to their importance in machine learning, finding meaningful intermediate representations has long been a primary research goal.
Disentangled representations can be interpreted more easily by humans and 
can for example be used to explain the reasoning of neural networks \cite{harradonCausalLearningExplanation2018}.

Among the longer known methods for extracting disentangled representations are principal component analysis (PCA) \cite{freyPrincipalComponentAnalysis1978, jolliffePrincipalComponentAnalysis2016}, independent component analysis \cite{hyvarinenIndependentComponentAnalysis2000}, and nonnegative matrix factorization \cite{berryAlgorithmsApplicationsApproximate2007}.
PCA is highly sensitive to outliers and noise in the data.
Therefore, more robust algorithms were proposed.
In \cite{sprechmannLearningRobustLowRank2012} already a small neural network was used as an encoder and the algorithm proposed in \cite{fengRobustPCAHighdimension2012} can deal with high-dimensional data.
Some robust PCA algorithms are provided with analytical performance guarantees \cite{xuRobustPCAOutlier2010, rahmaniCoherencePursuitFast2017, rahmaniOutlierDetectionData2019}.

A popular method for representation learning with deep networks is the variational autoencoder (VAE) \cite{kingmaAutoEncodingVariationalBayes2014}.
An important generalization of the method is the $ \beta $-VAE variant \cite{higginsBetaVAELearningBasic2017}, which improved the disentanglement capability \cite{fertigBetaVAEsCan2018}.
Later analysis added to the theoretical understanding of $ \beta $-VAE
\cite{burgessUnderstandingDisentanglingBeta2018, sikkaCloserLookDisentangling2019, kumarImplicitRegularizationBeta2020}.
Compared to standard autoencoders, VAEs map inputs to a distribution, instead of mapping them to a fixed vector.
This allows for additional regularization of the training to avoid overfitting and ensure good representations.
In $ \beta $-VAEs the trade-off between reconstruction quality and disentanglement can be fine-tuned by the hyperparameter $ \beta $.

Different regularization schemes have been suggested to improve the VAE method.
Among them are Wasserstein autoencoders \cite{tolstikhinWassersteinAutoEncoders2019, xiaoDisentangledRepresentationLearning2019}, attribute regularization \cite{patiAttributebasedRegularizationVAE2020} and relational regularization \cite{xuLearningAutoencodersRelational2020}.
Recently, a connection between VAEs and nonlinear independent component analysis was established \cite{khemakhemVariationalAutoencodersNonlinear2020} and then expanded \cite{sorrensonDisentanglementNonlinearICA2020}.

Besides VAEs, deep generative adversarial networks can be used to construct latent features \cite{sohnLearningStructuredOutput2015, chenInfoganInterpretableRepresentation2016, Makhzani16:AAE}.
Other works suggest centroid encoders \cite{ghoshSupervisedDimensionalityReduction2020} or conditional learning of Gaussian distributions \cite{sunConditionalGaussianDistribution2020} as alternatives to VAEs.
In \cite{kim_interpretability_2018} concept activation vectors are defined as being orthogonal to the decision boundary of a classifier.
Apart from deep learning, entirely new architectures, such as capsule networks \cite{sabourDynamicRoutingCapsules2017}, might be used to disassemble inputs. 
\par
While many different approaches for disentangling exist, the feasibility of the task is not clear yet and a better theoretical understanding is needed.
The disentangling performance is hard to quantify, which is only feasible with information about the latent ground truth \cite{eastwoodFrameworkQuantitativeEvaluation2018}.
Models that overly rely on single directions, single neurons in fully connected networks or single feature maps in CNNs, have the tendency to overfit \cite{morcosImportanceSingleDirections2018}.
According to \cite{locatelloChallengingCommonAssumptions2019}, unsupervised learning does not produce good disentangling and even small latent spaces do not reduce the sample complexity for simple tasks.
This is in direct contrast to newer findings that show a decreased sample complexity for more complex visual downstream tasks \cite{vansteenkisteAreDisentangledRepresentations2020}.
So far, it is unclear if disentangling improves the performance of machine learning tasks.

In order to be interpretable, latent disentangled representations need to be aligned with human understandable concepts.
In \cite{engstromAdversarialRobustnessPrior2019a} training with adversarial examples was used and the learned representations were shown to be more aligned with human perception.
For explainable AI, disentangling alone might not be enough to generate interpretable output and additional regularization could be needed.

\subsection{Pixel Attribution}\label{subsec:interpretability:pixel_attribution}
\sectionauthor{Stephanie Abrecht\textsuperscript{2}, Felix Brockherde\textsuperscript{10}, Toshika Srivastava\textsuperscript{12}}
The non-linearity and complexity of DNNs allow them to solve perception problems, like detecting a pedestrian, that cannot be specified in detail.  At the same time, the automatic extraction of features given in an input image and the mapping to the respective prediction is counterintuitive and incomprehensible for humans, which makes it hard to argue safety for a neural network-based perception task. Feature importance techniques are currently predominantly used to diagnose the causes of incorrect model behaviors \cite{Bhatt2020}. So-called \emph{attribution maps} are a visual technique to express the relationship between relevant pixels in the input image and the network's prediction. Regions in an image that contain relevant features are highlighted accordingly. 
Attribution approaches mostly map to one of three categories.
\par
Gradient-based and activation-based approaches (such as \cite{Selvaraju2016,Simonyan2014,Shrikumar2017,Shrikumar2019,Smilkov17,Bach2015,Springenberg2014,Montavon2015} amongst others) rely on the gradient of the prediction with respect to the input. Regions that were most relevant for the prediction are highlighted. Activation-based approaches relate the feature maps of the last convolutional layer to output classes. 
\par
Perturbation-based approaches \cite{Fong_2017,Zeiler2018,Zintgraf2017,Hoyer2019} suggest manipulating the input. If the prediction changes significantly, the input may hold a possible explanation at least.
\par
While gradient-based approaches are oftentimes faster in computation, perturbation-based approaches are much easier to interpret.
\par
As many studies have shown \cite{Sundararajan2017,Adebayo2018}, there is still a lot of research to be done before attribution methods are able to robustly provide explanations for model predictions, in particular erroneous behavior. One key difficulty is the lack of an agreed-upon definition of a good attribution map including important properties. Even between humans, it is hard to agree on what a good explanation is due to its subjective nature. This lack of ground truth makes it hard or even impossible to quantitatively evaluate an explanation method. Instead, this evaluation is done only implicitly. 
One typical way to do this is the axiomatic approach. Here a set of desiderata of an attribution method are defined, on which different attribution methods are then evaluated. Alternatively, different attribution methods may be compared by perturbing the input features starting with the ones deemed most important and measuring the drop in accuracy of the perturbed models. The best method will result into the greatest overall loss in accuracy as the number of inputs are omitted \cite{Ancona2017}.
Moreover, for gradient-based methods it is hard to assess if an unexpected attribution is caused by a poorly performing network or a poorly performing attribution method \cite{Fong_2017}. How to cope with negative evidence, \ie the object was predicted because a contrary clue in the input image was missing, is an open research question. Additionally, most methods were shown on classification tasks until now. It remains to be seen how they can be transferred to object detection and semantic segmentation tasks. In the case of perturbation-based methods, the high computation time and single-image analysis inhibit wide-spread application.

\subsection{Interpretable Proxies}\label{subsec:interpretability:interpretable_proxies}
\sectionauthor{Gesina Schwalbe\textsuperscript{3}}
Neural networks are capable of capturing complicated logical
(cor)relations. However, this knowledge is encoded on a
\emph{sub-symbolic} level in the form of learned weights and biases,
meaning that the reasoning behind the processing chain cannot be
directly read out or interpreted by humans \cite{cristea_neural_2001}.
To explain the sub-symbolic processing, one can either use attribution
methods (\cf \refsec{subsec:interpretability:pixel_attribution}), or lift
this sub-symbolic representation to a \emph{symbolic} one
\cite{gilpin_explaining_2018}, meaning a more interpretable one.
Interpretable proxies or surrogate models try to achieve the latter:
The DNN behavior is approximated by a model that
uses symbolic knowledge representations.
Symbolic representations can be
linear models like LIME \cite{ribeiro_why_2016} (proportionality),
decision trees (if-then-chains) \cite{gilpin_explaining_2018},
or loose sets of logical rules.
Logical connectors can simply be AND and OR but also more general
ones like at-least-M-of-N \cite{cristea_neural_2001}.
The \emph{expressiveness} of an approach refers to the logic that is
used: Boolean-only versus first-order logic, and binary versus fuzzy
logic truth values \cite{tickle_truth_1998}.
Other than attribution methods (\cf \refsec{subsec:interpretability:pixel_attribution}), these
representations can capture combinations of features and (spatial)
relations of objects and attributes. As an example consider
\enquote{eyes are closed} as explanation for \enquote{person asleep}:
Attribution methods only could mark the location of the eyes, dismissing
the relations of the attributes \cite{rabold_explaining_2018}.
All mentioned surrogate model types (linear, set of rules) require
interpretable input features in order to be interpretable themselves.
These features must either be directly obtained from the DNN input or
(intermediate) output, or automatically be extracted from the DNN
representation.
Examples for extraction are the super-pixeling used in LIME for input
feature detection, or concept activation vectors
\cite{kim_interpretability_2018} for DNN representation decoding.
\par
Quality criteria and goals for interpretable proxies are \cite{tickle_truth_1998}:
\emph{accuracy} of the standalone surrogate model on unseen examples,
\emph{fidelity} of the approximation by the proxy,
\emph{consistency} with respect to different training sessions, and
\emph{comprehensibility} measured by the complexity of the rule set
(number of rules, number of hierarchical dependencies).
The criteria are usually in conflict and need to be balanced:
Better accuracy may require a more complex, thus less expressive sets of rules.
\par
Approaches for interpretable proxies differ in the validity range of
the representations:
Some aim for surrogates that are only valid \emph{locally} around
specific samples, like in LIME~\cite{ribeiro_why_2016} or in
\cite{rabold_explaining_2018} via inductive logic programming.
Other approaches try to more \emph{globally} approximate aspects of
the model behavior.
Another categorization is defined by whether full access
(\emph{white-box}), some access (\emph{gray-box}), or no access
(\emph{black-box}) to the DNN internals is needed.
One can further differentiate between \emph{post-hoc} approaches that
are applied to a trained model, and approaches that try to integrate
or \emph{enforce symbolic representations} during training.
Post-hoc methods cover the wide field of rule extraction techniques
for DNNs. The reader may refer to
\cite{hailesilassie_rule_2016,augasta_rule_2012}.
Most white- and gray-box methods try to turn the DNN connections into
if-then rules that are then simplified, like done in
DeepRED~\cite{zilke_deepred_2016}.
A black-box example is validity interval
analysis~\cite{thrun_extracting_1995}, which refines or
generalizes rules on input intervals, either starting from one sample
or a general set of rules.
Enforcement of symbolic representations can be achieved by enforcing
an output structure that provides insights to the decision logic, such as
textual explanations, or a rich output structure allowing investigation of correlations
\cite{xiao_unified_2018}.
An older discipline for enforcing symbolic representations
is the field of neural-symbolic learning
\cite{svatos_revisiting_2019}. The idea is based on a hybrid learning
cycle 
in which a symbolic learner and a DNN iteratively update each other via
rule insertion and extraction.
\par
The comprehensibility of global surrogate models suffers from the
complexity and size of concurrent DNNs. Thus, stronger rule
simplification methods are required \cite{hailesilassie_rule_2016}.
The alternative direction of local approximations mostly
concentrates on linear models instead of more expressive rules \cite{thrun_extracting_1995,rabold_explaining_2018}.
Furthermore, balancing of the quality objectives is hard since
available indicators for interpretability may not be ideal.
And lastly, applicability is heavily infringed by the requirement of
interpretable input features. These are usually not readily available
from input (often pixel-level) or DNN output. Supervised extraction
approaches vary in their fidelity, and unsupervised ones do not
guarantee to yield meaningful or interpretable results, respectively, such
as the super-pixel clusters of LIME.

\section{Uncertainty} \label{sec:uncertainty} 
\sectionauthor{Michael Mock\textsuperscript{1}}
Uncertainty refers to the view that a neural network is not conceived as a deterministic function but as a probabilistic function or estimator, delivering a random distribution for each input point. Ideally, the mean value of the distribution should  be as close as possible to the ground truth value of the function being approximated by the neural network and the uncertainty of the neural network refers to its variance when being considered as a random variable, thus allowing to derive a confidence with respect to the mean value.  Regarding safety, the variance may lead to estimations about the confidence associated with a specific network output and opens the option for discarding network outputs with insufficient confidence. 
\par
There are roughly two broad approaches for training neural networks as probabilistic functions: Parametric approaches \cite{kendall2017uncertainties} and Bayesian neural networks on the one hand, such as \cite{weightUncertaintyInNN}, where the transitions along the network edges are modeled as probability distributions, and ensemble-based approaches on the other hand \cite{lakshminarayanan2017simple,snoek2019can}, where multiple networks are trained and considered as samples of a common output distribution. Apart from training as probabilistic function, uncertainty measures have been derived from single, standard neural networks by post-processing on the trained network logits, leading for example to calibration measures (\cf \eg \cite{snoek2019can}). 

\subsection{Generative Models} \label{subsec:uncertainty:generative_models}
\sectionauthor{Sebastian Wirkert\textsuperscript{6}, Tim Wirtz\textsuperscript{1}}
\emph{Generative models} belong to the class of unsupervised machine learning models. From a theoretical perspective, these are particularly interesting, because they offer a way to analyze and model the density of data. Given a finite data set $\mathcal{D}$ independently distributed according to some distribution $p(x)$, generative models aim to estimate or enable sampling from the underlying density $p(x)$ in a model $q_\theta(x)$. The resulting model can be used for data indexing~\cite{westerveld2004using}, data retrieval~\cite{miotto2011generative}, for visual recognition~\cite{Krizhevsky2012}, speech recognition and generation~\cite{hinton2012deep}, language processing~\cite{klein2003fast,cotterell2017probabilistic} and robotics~\cite{thrun2002robotic}.
Following \cite{oussidi2018deep}, we can group generative models into two main classes:
\begin{itemize}
    \item Cost function-based models such as autoencoder~\cite{kingmaAutoEncodingVariationalBayes2014,doersch2016tutorial}, deep belief networks~\cite{hinton2009deep} and generative adversarial networks~\cite{goodfellow2016nips,radford2015unsupervised,goodfellow2014generative}.
    \item Energy-based models~\cite{lecun2006tutorial,salakhutdinov2009deep}, where the joint probability density is modeled by an energy function.
\end{itemize}
Beside these \emph{deep learning} approaches, generative models have been studied in machine learning in general for quite some time (\cf \cite{wertz1978statistical,jones1996brief,fryer1977review,silverman1986density,gramacki2018nonparametric,scott2015multivariate,sheather2004density}). A very prominent example of generative networks are Gaussian processes~\cite{rasmussen2003gaussian,mackay1998introduction,williams1996gaussian,williams1998bayesian} and their deep learning extensions~\cite{damianou2013deep,bui2016deep} as generative models. 
\par
An example of a generative model being employed for image segmentation uncertainty estimation is the probabilistic U-Net \cite{kohl2018probabilistic}. Here a variational autoencoder (VAE) conditioned on the image is trained to model uncertainties. Samples from the VAE are fed into a segmentation U-Net which can thus give different results for the same image. This was tested in context of medical images, where inter-rater disagreements lead to uncertain segmentation results and Cityscapes segmentation. For the Cityscapes segmentation the investigated use case was label ambiguity (\eg is a BMW X7 a car or a van) using artificially created, controlled ambiguities. Results showed that the probabilistic U-Net could reproduce the segmentation ambiguity modes more reliably than competing methods such as a dropout U-Net which is based on techniques elaborated in the next section.

\subsection{Monte-Carlo Dropout} \label{subsec:uncertainty:mc_dropout} 
\sectionauthor{Joachim Sicking\textsuperscript{1}}
A widely used technique to estimate model uncertainty is Monte-Carlo (MC) dropout \cite{gal2016dropout}, that offers a Bayesian motivation, conceptual simplicity and scalability to application-size networks. This combination distinguishes MC dropout from competing Bayesian neural network (BNN) approximations (like \cite{weightUncertaintyInNN},\cite{laplaceApproximation}, see \refsec{subsec:uncertainty:bayesian_neural_networks}). However, these approaches and MC dropout share the same goal: to equip neural networks with a \emph{self-assessment} mechanism that detects unknown input concepts and thus potential model insufficiencies.

On a technical level, MC dropout assumes prior distributions on network activations, usually independent and identically distributed (i.i.d.) Bernoulli distributions. Model training with iteratively drawn Bernoulli samples, the so-called \emph{dropout masks}, then yields a data-conditioned posterior distribution within the chosen parametric family. It is interesting to note that this training scheme was used earlier---independent from an uncertainty context---for better model generalization \cite{srivastava2014dropout}. At inference, sampling provides estimates of the input-dependent output distributions. The spread of these distributions is then interpreted as the prediction uncertainty that originates from limited knowledge of model parameters. Borrowing ‘frequentist’ terms, MC dropout can be considered as an implicit network ensemble, \ie as a set of networks that share (most of) their parameters.

In practice, MC dropout requires only a minor modification of the optimization objective during training and multiple, trivially parallelizable forward passes during inference. The loss modification is largely agnostic to network architecture and does not cause substantial overhead. This is in contrast to the sampling-based inference that increases the computational effort massively---by estimated factors of 20-100 compared to networks without MC dropout. A common practice is therefore the use of \emph{last-layer dropout} \cite{snoek2019can} that reduces computational overhead to estimated factors of 2-10. Alternatively, analytical moment propagation allows sampling-free MC-dropout inference at the price of additional approximations (\eg\cite{postels2019sampling}). Further extensions of MC dropout target the integration of data-inherent (aleatoric) uncertainty \cite{kendall2017uncertainties} and tuned performance by learning layer-specific dropout rate using concrete relaxations \cite{gal2017concrete}.

The quality of MC-dropout uncertainties is typically evaluated using negative log-likelihood (NLL), expected calibration error (ECE) and its variants (\cf \refsec{subsec:uncertainty:confidence_calibration}) and by considering correlations between uncertainty estimates and model errors (\eg AUSE \cite{ilg2018uncertainty}). Moreover, it is common to study how useful uncertainty estimates are for solving auxiliary tasks like out-of-distribution classification \cite{lakshminarayanan2017simple} or robustness \wrt adversarial attacks. 

MC dropout is a working horse of safe ML, being used with various networks and for a multitude of applications (\eg\cite{bhattacharyya2018long}). However, several authors pointed out shortcomings and limitations of the method: MC dropout bears the risk of over-confident false predictions (\cite{osband2016risk}), offers less diverse uncertainty estimates compared to (the equally simple and scalable) deep ensembles (\cite{lakshminarayanan2017simple}, see \refsec{subsec:aggregation:ensemble_methods}) and provides only rudimentary approximations of true posteriors. 

Relaxing these modelling assumptions and strengthening the Bayesian motivation of MC dropout is therefore an important research avenue. Further directions for future work are the development of \emph{semantic uncertainty mechanisms} (\eg \cite{kohl2018probabilistic}), improved local uncertainty calibrations and a better understanding of the outlined sampling-free schemes to uncertainty estimation.
\subsection{Bayesian Neural Networks}\label{subsec:uncertainty:bayesian_neural_networks}
\sectionauthor{Maram Akila\textsuperscript{1}}
As the name suggests, Bayesian neural networks (BNNs) are inspired by a Bayesian interpretation of probability (for an introduction \cf \cite{mcKayBook}).
In essence, it rests on Bayes' theorem,
\begin{equation}
	p(x|y) p(y) = p(x,y) = p(y | x) p(x)
	\quad \Rightarrow \quad
	p(x| y) = \frac{ p(y | x) p(x)}{ p(y) }
	\,,
	\label{eq:bnn:bayes}
\end{equation}
stating that the conditional probability density function (PDF) $p(x|y)$ for $x$ given $y$ may be expressed in terms of the inverted conditional PDF $p(y | x)$.
For machine learning, where one intends to make predictions $y$ for unknown $x$ given some training data $\mathcal{D}$, this can be reformulated into
\begin{equation}
	y = \operatorname{NN}(x|W)
	\quad\text{with}\quad
	 p(W | \mathcal{D}) = \frac{p(\mathcal{D} | W) p(W)}{p(\mathcal{D})}
	\,.
	\label{eq:bnn:NNpdf}
\end{equation}
Therein NN denotes a conventional (deep) neural network (DNN) with model parameters $W$, \eg the set of weights and biases.
In contrast to a regular DNN, the weights are given in terms of a probability distribution $ p(W | \mathcal{D}) $ turning also the output $y$ of a BNN into a distribution.
This allows to study the mean $\mu=\langle y^1 \rangle$ of the DNN for a given $x$ as well as higher moments of the distribution, typically the resulting variance $\sigma^2 = \langle (y-\mu)^2 \rangle$ is of interest, where
\begin{equation}
	\left\langle y^k \right\rangle = \int\!
	\operatorname{NN}(x|W)^k p(W|\mathcal{D})\,\mathrm{d}W\,.
	\label{eq:bnn:moments}
\end{equation}
While $\mu$ yields the output of the network, the variance $\sigma^2$ is a measure for the uncertainty of the model for the prediction at the given point.
Central to this approach is the probability of the data given the model, here denoted by $p(\mathcal{D} | W) $, as it is the key component connecting model and training data.
Typically, the prior distribution $p(\mathcal{D})$ is ``ignored'' as it only appears as a normalization constant within the averages, see \eqref{eq:bnn:moments}.
In the cases where the data $\mathcal{D}$ is itself a distribution due to inherent uncertainty, \ie presence of aleatoric risk \cite{kendall2017uncertainties}, such a concept seems natural.
However, Bayesian approaches are also applicable for all other cases.
In those, loosely speaking, the likelihood of $W$ is determined via the chosen loss function (for the connection between the two concepts \cf \cite{bishop2006pattern}).
\par
On this general level, Bayesian approaches are broadly accepted and also find use for many other model classes besides neural networks.
However, the loss surfaces of DNNs are known for their high dimensionality and strong non-convexity.
Typically, there are abundant parameter combinations $W$ that lead to (almost) equally good approximations to the training data $\mathcal{D}$ with respect to a chosen loss.
This makes an evaluation of $p(W|\mathcal{D})$ for DNNs close to impossible in full generality.
At least no (exact) solutions for this case exist at the moment.
\par
Finding suitable approximations to the posterior distribution $p(W|\mathcal{D})$  is an ongoing challenge for the construction of BNNs.
At this point we only summarize two major research directions in the field.
One approach is to assume that the distribution factorizes.
While the full solution would be a joined distribution implying correlations between different weights (etc.), possibly even across layers, this approximation takes each element $w_i$ of $W$ to be independent form the others.
Although this is a strong assumption, it is often made, in this case  parameters for the respective distributions of each element can be learned via training (\cf \cite{weightUncertaintyInNN}).
The second class of approaches focuses on the region of the loss surface around the minimum chosen for the DNN.
As discussed, the loss relates to the likelihood and quantities such as the curvature at the minimum, therefore directly connected to the distribution of $W$.
Unfortunately, already using this type of quantities requires further approximations \cite{laplaceApproximation}.
Alternatively, the convergence of the training process may be altered to sample networks close to the minimum \cite{langevinDynamics}.
While this approach contains information about correlations among the $w_i$, it is usually restricted to a specific minimum.
For a non-Bayesian ansatz taking into account several minima, see deep ensembles in \refsec{subsec:aggregation:ensemble_methods}.
BNNs also touch other concepts such as MC dropout (\cf \refsec{subsec:uncertainty:mc_dropout} or \cite{gal2016dropout}), or prior networks, which are based on a Bayesian interpretation but use conventional DNNs with an additional (learned) $\sigma$ output \cite{uncertatintyViaPrior}.

\subsection{Uncertainty Metrics for DNNs in Frequentist Inference}\label{subsec:uncertainty:gradient_based_uncertainty}
\sectionauthor{Matthias Rottmann\textsuperscript{5}}
Classical uncertainty quantification methods in frequentist inference are mostly based on the outputs of statistical models. Their uncertainty is quantified and assessed for instance via dispersion measures in classification (such as entropy, probability margin or variation ratio), or confidence intervals in regression. However, the nature of DNN architectures \cite{Redmon2015,chen2017deeplab} and the cutting edge applications tackled by those (\eg semantic segmentation, \cf \cite{Cordts2016}) open the way towards more elaborate uncertainty quantification methods. Besides the mentioned classical approaches, intermediate feature representations within a DNN (\cf \cite{okamoto2020outofdistribution,CHENG201921}) or gradients according to self-affirmation that represent re-learning stress (see \cite{Oberdiek2018}) reveal additional information. In addition, in case of semantic segmentation, the geometry of a prediction may give access to further information, cf.\ \cite{MetaSeg,Schubert2019,Maag2019}. By the computation of statistics of those quantities as well as low-dimensional representations thereof, we obtain more elaborate uncertainty quantification methods specifically designed for DNNs that can help us to detect misclassifications and out-of-distribution objects (\cf \cite{hendrycks2017baseline}). 
\par
Features gripped during a fordward pass of a data point $x$ through a DNN $f$ can be considered layer-wise, \ie $f^{(\ell)}(x)$ after the $\ell$-th layer. These can be translated into a handful of quantities per layer \cite{okamoto2020outofdistribution} or further processed by another DNN that aims that detecting errors \cite{CHENG201921}. While in particular \cite{okamoto2020outofdistribution} presents a proof of concept on small scale classification problems, their applicability to large scale datasets and problems such as semantic segmentation and object detection remain open.
\par
The development for gradient-based uncertainty quantification methods \cite{Oberdiek2018} is guided by one central question:
If the present prediction was true, how much re-learning would this require.
The corresponding hypothesis is that wrong predictions would be more in conflict with the knowledge encoded in the deep neural network than correct ones, therefore causing increased re-learning stress.
Given a predicted class
\begin{equation}
\hat{y} = \mathop{\mathrm{arg\,max}}_{y} f_y(x) 
\end{equation}
we compute the gradient of layer $\ell$ corresponding to the predicted label. That is, given a loss function $\mathcal{L}$, we compute
\begin{equation}
\nabla_{w_\ell} \mathcal{L}( \hat{y}, x, w ) 
\end{equation}
via backpropagation. The obtained quantities can be treated similarly to the case of forward pass features. While this concept seems to be prohibitively expensive for semantic segmentation (at least when calculating gradients for each pixel of $\hat{y}$), its applicability to object detection might be feasible, in particular with respect to offline applications. Gradients are also of special interest in active learning with query by \emph{expected model change} (\cf \cite{Settles2010}).
\par
In the context of semantic segmentation, geometrical information on segments shapes as well as neighborhood relations of predicted segments can be taken into account along side with dispersion measures. It has been demonstrated \cite{MetaSeg,Schubert2019,Maag2019} that the detection of errors in an in-distribution setting strongly benefits from geometrical information. Recently, this has also been considered in scenarios under moderate domain shift \cite{oberdiek2020detection}. However, its applicability to out-of-distribution examples and to other sensors than the camera remains subject to further research.
\subsection{Markov Random Fields}\label{subsec:uncertainty:markov_random_fields}
\sectionauthor{Seyed Eghbal Ghobadi\textsuperscript{8}, Ahmed Hammam\textsuperscript{8}}
Although deep neural networks are currently the state of the art for almost all computer vision tasks, Markov random fields (MRF) remain one of the fundamental techniques used for many computer vision tasks, specifically image segmentation \cite{krahenbuhl2011efficient},\cite{li2009sar}. MRFs hold its power in the essence of being able to model dependencies between pixels in an image. With the use of energy functions, MRFs integrate pixels into models relating between unary and pairwise pixels together \cite{wang2013markov}. 
Given the model, MRFs are used to infer the optimal configuration yielding the lowest energy using mainly maximum a posteriori (MAP) techniques. Several MAP inference approaches are used to yield the optimal configuration such as graph cuts \cite{kohli2008simultaneous} and belief propagation algorithms \cite{felzenszwalb2010dynamic}. However, as with neural networks, MAP inference techniques result in deterministic point estimates of the optimal configuration without any sense of uncertainty in the output. To obtain uncertainties on results from MRFs, most of the work is directed towards modelling MRFs with Gaussian distributions. Getting uncertainties from MRFs with Gaussian distributions is possible by two typical methods: Either approximate models are inferred to the posterior, from which sampling is easy or the variances can be estimated analytically, or approximate sampling from the posterior is used.
Approximate models include those inferred using variational Bayesian (VB) methods, like mean-field approximations, and using Gaussian process (GP) models enforcing a simplified prior model \cite{le2016sampling}, \cite{bishop2006pattern}. Examples of approximate sampling methods include traditional Markov chain Monte Carlo (MCMC) methods like Gibbs sampling \cite{4767596}. Some recent theoretical advances propose the perturb-and-MAP framework and a Gumbel perturbation model (GPM) \cite{papandreou2011perturb},\cite{hazan2013sampling} to exactly sample from MRF distributions. Another line of work has also been proposed, where  MAP inference techniques are used to estimate the probability of the network output. With the use of graph cuts, \cite{kohli2008measuring} try to estimate uncertainty using the min-marginals associated with the label assignments of a random field. Here, the work by Kohli and Torr \cite{kohli2008measuring} was extended to show how this approach can be extended to techniques other than graph cuts \cite{tarlow2012revisiting} or compute uncertainties on multi-label marginal distributions \cite{summa2017visualizing}.
\par
A current research direction is the incorporation of MRFs with deep neural networks, along with providing uncertainties on the output \cite{schwing2015fully, chen2017deeplab}. This can also be extended to other forms of neural networks such as recurrent neural networks to provide uncertainties on segmentation of streams of videos with extending dependencies of pixels to previous frames \cite{liu2017deep}, \cite{zheng2015conditional}.

\subsection{Confidence Calibration} \label{subsec:uncertainty:confidence_calibration} 
\sectionauthor{Fabian Küppers\textsuperscript{9}, Anselm Haselhoff\textsuperscript{9}}
Neural network classifiers output a label $\hat{Y} \in \mathcal{Y}$ on a given input $X \in \mathcal{X}$ with an associated confidence $\hat{P}$. This confidence can be interpreted as a probability of correctness that the predicted label matches the ground truth label $Y \in \mathcal{Y}$. Therefore, these probabilities should reflect the ”self-confidence” of the system. If the empirical accuracy for any confidence level matches the predicted confidence, a model is called \emph{well calibrated}.
Therefore, a \emph{classification} model is perfectly calibrated if 
\begin{align}
\label{eq:classification_calibration}
& \underbrace{\mathbb{P}(\hat{Y} = Y | \hat{P} = p)}_{\text{accuracy given } p} = \underbrace{p}_{\text{confidence}} \quad \forall p \in [0,1]
\end{align}
is fullfilled \cite{Guo2017}. For example, assume 100 predictions with confidence values of 0.9. We call the model well calibrated if 90 out of these 100 predictions are actually correct. However, recent work has shown that modern neural networks tend to be too overconfident in their predictions \cite{Guo2017}. The deviation of a model to the perfect calibration can be measured by the expected calibration error (ECE) \cite{Naeini2015}. 
It is possible to recalibrate models as a post-processing step after classification. One way to get a calibration mapping is to group all predictions into several bins by their confidence. Using such a binning scheme, it is possible to compute the empirical accuracy for certain confidence levels, as it is known for a long time already in reconstructing confidence outputs for Viterbi decoding \cite{huber_sova}. Common methods are histogram binning \cite{Zadrozny2001}, isotonic regression \cite{Zadrozny2002} or more advanced methods like Bayesian binning into quantiles (BBQ) \cite{Naeini2015} and ensembles of near-isotonic regression (ENIR) \cite{Naeini2015a}. Another way to get a calibration mapping is to use scaling methods based on logistic regression like Platt scaling \cite{Platt1999}, temperature scaling \cite{Guo2017} and beta calibration \cite{Kull2017}.
\par
In the setting of probabilistic regression, a model is calibrated if, \eg 95\% of the true target values are below or equal to a credible level of 95\% (so called \emph{quantile-calibrated regression}) \cite{Gneiting2007, Kuleshov2018, Song2019}. A regression model is usually calibrated by fine-tuning its predicted CDF in a post-processing step to match the empirical frequency. Common approaches utilize isotonic regression \cite{Kuleshov2018}, logistic and beta calibration \cite{Song2018}, as well as Gaussian process models \cite{Song2018, Song2019} to build a calibration mapping. In contrast to quantile-calibrated regression, \cite{Song2019} have recently introduced the concept of distribution calibration, where calibration is applied on a distribution level and naturally leads to calibrated quantiles.
\par
Recent work has shown that miscalibration in the scope of \emph{object detection} also depends on the position and scale of a detected object \cite{Kueppers2020}. The additional box regression output is denoted by $\hat{R}$ with $J$ as the size of the used box encoding.
Furthermore, if we have no knowledge about all anchors of a model (which is a common case in many applications), it is not possible to determine the accuracy. Therefore, K\"{u}ppers \etal~\cite{Kueppers2020} use the precision as a surrogate for accuracy and propose that an \emph{object detection model} is perfectly calibrated if
\begin{align}
\label{eq:detection_calibration}
& \underbrace{\mathbb{P}(M=1 | \hat{P} = p, \hat{Y} = y, \hat{R} = r)}_{\text{precision given } p, y, r} = \underbrace{p}_{\text{confidence}} \quad \forall p \in [0,1], y \in \mathcal{Y}, r \in \mathbb{R}^J 
\end{align}
is fulfilled, where $M=1$ denotes a correct prediction that matches a ground-truth object with a chosen IoU threshold and $M=0$ denotes a mismatch, respectively. The authors propose the detection-expected calibration error (D-ECE) as the extension of the ECE to object detection tasks in order to measure miscalibration also by means of the position and scale of detected objects. Other approaches try to fine-tune the regression output in order to obtain more reliable object proposals \cite{Jiang2018, Rezatofighi2019} or to add a regularization term to the training objective such that training yields models that are both well-performing and well-calibrated \cite{Pereyra2017, Seo2019}.

\section{Aggregation}\label{sec:aggregation}
\sectionauthor{Maram Akila\textsuperscript{1}}
From a high-level perspective, a \emph{neural network} is based on processing inputs and coming to some output conclusion, \eg mapping incoming image data onto class labels.
Aggregation or collection of non-independent information on either the input or output side of this network function can be used as a tool to leverage its performance and reliability.
Starting with the input, any additional ``dimension'' to add data can be of use.
For example, in the context of autonomous vehicles this might be input from any further sensor measuring the same scene as the original one, \eg stereo cameras or LiDAR.
Combining those sensor sets for prediction is commonly referred to as \emph{sensor fusion} \cite{agg_lidar}.
Staying with the example, the scene will be monitored consecutively providing a whole (temporally ordered)  stream of input information.
This may be used either by adjusting the network for this kind of input 
\cite{agg_dnnApproach2} 
or in terms of a post-processing step, in which the predictions are aggregated by some measure of temporal consistency. 
\par
Another more implicit form of aggregation is training the neural network on several ``independent'' tasks, \eg segmentation and depth regression. Although the individual task is executed on the same input, the overall performance can still benefit from the correlation among all given tasks. We refer to the discussion on \emph{multi-task networks} in \refsec{subsec:architecture:multi_task_networks}.
By extension, solving the same task in multiple different ways, can be beneficial for performance and provide a measure of redundancy.
In this survey, we focus on single-task systems and discuss \emph{ensemble methods} in the next section and the use of \emph{temporal consistency} in the one thereafter.

\subsection{Ensemble Methods}\label{subsec:aggregation:ensemble_methods}
\sectionauthor{Joachim Sicking\textsuperscript{1}}
Training a neural network is optimizing its parameters to fit a given training data set. The commonly used gradient-based optimization schemes cause convergence in a ‘nearby’ local minimum. As the loss landscapes of neural networks are notoriously non-convex \cite{choromanska2015loss}, various locally optimal model parameter sets exist. These local optima differ in the degree of optimality (“deepness”), qualitative characteristics (“optimal for different parts of the training data”) and their generalizability to unseen data (commonly referred to by the geometrical terms of “sharpness” and “flatness” of minima \cite{keskar2016large}). 

A single trained network corresponds to one local minimum of such a loss landscape and thus captures only a small part of a potentially diverse set of solutions. \emph{Network ensembles} are collections of models and therefore better suited to reflect this multi-modality. Various modelling choices shape a loss landscape: the selected model class and its meta-parameters (like architecture and layer width), the training data and the optimization objective. Accordingly, approaches to diversify ensemble components range from combinations of different model classes over varying training data (bagging) to methods that train and weight ensemble components to make up for the flaws of other ensemble members (boosting) \cite{bishop2006pattern}.

Given the millions of parameters of application-size networks, ensembles of NNs are resource-demanding \wrt computational load, storage and runtime during training and inference. This complexity increases linearly with ensemble size for naïve ensembling. Several approaches were put forward to reduce some dimensions of this complexity: \emph{snapshot ensembles} \cite{huang2017snapshot} require only one model optimization with a cyclical learning-rate schedule—leading to an optimized training runtime.  The resulting training trajectory passes through several local minima. The corresponding models compose the ensemble. On the contrary, \emph{model distillation} \cite{hinton2015distilling} tackles runtime at inference. They ‘squeeze’ a NN ensemble into a single model that is optimized to capture the gist of the model set. However, such a compression goes along with reduced performance compared to the original ensemble.

Several hybrids of single model and model ensemble exist: Multi-head networks \cite{arik2018fast} share a backbone network that provides inputs to multiple prediction networks. Another variant are mixture-of-expert models that utilize a gating network to assign inputs to specialized expert networks \cite{shazeer2017outrageously}. Multi-task networks (\cf \refsec{subsec:architecture:multi_task_networks}) and Bayesian approximations of NNs (\cf \refsec{subsec:uncertainty:bayesian_neural_networks} and \refsec{subsec:uncertainty:mc_dropout}) can be seen as implicit ensembles. 

NN ensembles (or deep ensembles) are not only used to boost model quality. They pose the frequentist's approach to estimating NN uncertainties and are state-of-the-art in this regard \cite{lakshminarayanan2017simple,snoek2019can}. The emergent field of federated learning is concerned with the integration of decentrally trained ensemble components \cite{mcmahan2016communication} and safety-relevant applications of ensembling range from autonomous driving \cite{zhang2012reliable} to medical diagnostics \cite{rasti2017macular}. 
Taking this safe-ML perspective, promising research directions comprise a more principled and efficient composition of model ensembles, \eg by application-driven diversification, as well as improved techniques to miniaturize ensembles, \eg by gaining a better understanding of methods like distillation. In the long run, better designed, more powerful learning systems might partially reduce the need for combining weaker models in a network ensemble.
\subsection{Temporal Consistency}\label{subsec:aggregation:temporal_consistency}
\sectionauthor{Timo Sämann\textsuperscript{4}}
The focus of previous DNN development for semantic segmentation has been on single image prediction. This means that the final and intermediate results of the DNN are discarded after each image. However, the application of a computer vision model often involves the processing of images in a sequence, \ie there is a temporal consistency in the image content between consecutive frames (for a metric, \cf, \eg \cite{Varghese2020}). This consistency has been exploited in previous work to increase quality and reduce computing effort. Furthermore, this approach offers the potential to improve the robustness of DNN prediction by incorporating this consistency as a-priori knowledge into DNN development. The relevant work in the field of video prediction can be divided in two major approaches: 

\begin{enumerate}
\item DNNs are specially designed for video prediction. This usually requires training from scratch and the availability of training data in a sequence.

\item A transformation from single prediction DNNs to video prediction DNNs takes place. Usually no training is required, \ie the existing weights of the model can be used unaltered.
\end{enumerate}

The first set of approaches often involves \emph{conditional random fields} (CRF) and its variants. 
CRFs are known for their use as postprocessing step in the prediction of semantic segmentation, in which their parameters are learned separately or jointly with the DNN~\cite{zheng2015conditional}. 
Another way to use spatiotemporal features is to include 3D convolutions, which add an additional dimension to the conventional 2D convolutional layer. Tran \etal\cite{tran2015learning} use 3D convolution layers for video recognition tasks such as action and object recognition.     
One further approach to use spatial and temporal characteristics of the input data is to integrate \emph{long short-term memory} (LSTM)~\cite{hochreiter1997long}, a variant of the \emph{recurrent neural network} (RNN). Fayyaz \etal \cite{fayyaz2016stfcn} integrate LSTM layers between the encoder and decoder of their convolutional neural network for semantic segmentation. The significantly higher GPU memory requirements and computational effort are a disadvantage of this method. More recently, Nilsson and Sminchisescu \cite{nilsson2018semantic} deployed \emph{gated recurrent units}, which generally requires significantly less memory.
An approach to improve temporal consistency of automatic speech recognition outputs is known as a posterior-in-posterior-out (PIPO) LSTM ``sequence enhancer'', a postfilter which could be applicable to video processing as well \cite{lohrenz2019}.
A disadvantage of the described methods is that sequential data for training must be available, which may be limited or show a lack of diversity. 
\par
The second class of approaches has the advantage that it is model-independent most of the time. Shelhamer \etal\cite{shelhamer2016clockwork} found that the deep feature maps within the network change only slightly with temporal changes in video content.
Accordingly, \cite{gadde2017semantic} calculate the optical flow of the input images from time steps $t_0$ and $t_{-1}$ and convert it into the so-called \emph{transform flow} which is used to transform the feature maps of the time step $t_{-1}$ so that an aligned representation to the feature map $t_0$ is achieved. S{\"a}mann \etal\cite{samann2019robust} use a confidence-based combination of feature maps from previous time steps based on the calculated optical flow.

\section{Verification}\label{sec:verification}
\sectionauthor{Gesina Schwalbe\textsuperscript{3}}
Verification and validation is an integral part of the safety
assurance for any safety critical systems. As of the functional safety
standard for automotive systems \cite{iso/tc_22/sc_32_iso_2018},
\emph{verification} means to determine whether given requirements are
met \cite[3.180]{iso/tc_22/sc_32_iso_2018}, such as performance
goals.
\emph{Validation} on the other side tries to assess whether the given
requirements are sufficient and adequate to guarantee safety
\cite[3.148]{iso/tc_22/sc_32_iso_2018}, \eg whether certain types of
failures or interactions simply were overlooked. The latter is usually
achieved via extensive testing in real operation conditions of the
integrated product. This differs from the notion of validation used in
the machine learning community in which it usually refers to simple
performance tests on a selected dataset.
In this section, we want to concentrate on general verification
aspects for deep neural networks.

Verification as in the safety domain encompasses (manual) inspection
and analysis activities, and testing. However, the contribution
of single processing steps within a neural network to the final
behavior can hardly be assessed manually (compare to the problem of
interpretability in \refsec{sec:interpretability}). Therefore, we here will concentrate on different
approaches to verification testing.
Section~\ref{subsec:verification:formal_testing} covers approaches
to systematic test data selection. While the suggested methods assume
full access to the model internals for coverage measurement, this is
not in all cases available. Therefore,
\refsec{subsec:verification:black_box_methods} highlights
assessment approaches that consider the DNN as a black-box component.
Also, the topic of verification activity during operation with the
help of observers is discussed.

\subsection{Formal Testing}\label{subsec:verification:formal_testing}
\sectionauthor{Christian Heinzemann\textsuperscript{2}, Gesina Schwalbe\textsuperscript{3}, Matthias Woehrle\textsuperscript{2}}
Formal testing refers to testing methods that include formalized and
formally verifiable steps, \eg for test data acquisition, or for
verification in the local vicinity of test samples.
For image data, local testing around valid samples is usually more
practical than fully formal verification: (Safety) properties are not
expected to hold on the complete input space but only on the much
smaller unknown lower-dimensional manifold of real images \cite{woehrle_open_2019}.
Sources of such samples can be real ones or generated ones using an
input space formalization or a trained generative model.
\par
Coverage criteria for the data samples are commonly used for two purposes:
(a) deciding when to stop testing or
(b) identifying missing tests.
For CNNs, there are at least three different approaches towards
coverage:
(1) approaches that establish coverage based on a model with semantic
features of the input space~\cite{GHHW20},
(2) approaches trying to semantically cover the latent feature space
of neural network or a proxy network (\eg an
autoencoder)~\cite{schwalbe_survey_2020}, and
(3) approaches trying to cover neurons and their interactions,
inspired by classical software white-box
analysis~\cite{pei:2017,sun2019structural}.
\par
Typical types of properties to verify are 
simple test performance,
local stability (robustness),
a specific structure of the latent spaces like
embedding of semantic concepts \cite{schwalbe_concept_2020},
and more complex logical constraints on inputs and outputs,
which can be used for testing when fuzzified \cite{roychowdhury_image_2018}.
Most of these properties require in-depth semantic information about
the DNN inner workings, which is often only available via interpreting
intermediate representations \cite{kim_interpretability_2018},
or interpretable proxies / surrogates (\cf \refsec{subsec:interpretability:interpretable_proxies}), which do not guarantee fidelity.
\par
There exist different testing and formal verification methods from
classical software engineering that have already been applied to CNNs.
\emph{Differential testing} as used by
DeepXPlore~\cite{pei:2017} trains $n$ different CNNs for
the same task using independent data sets and compares the individual
prediction results on a test set. This allows to identify
inconsistencies between the CNNs but no common weak spots.
\emph{Data augmentation} techniques start from a given data set and
generate additional transformed data. \emph{Generic data augmentation}
for images like rotations and translation are state-of-the-art for
training but may also be used for testing.
\emph{Concolic testing} approaches incrementally grow test suites with
respect to a coverage model to finally achieve completeness. Sun \etal \cite{sun_concolic_2018} use an adversarial input model based on
some norm (\cf \refsec{subsec:adverarial_attacks:adversarial_attacks_defenses}),
\eg an $L_p$-norm, for generating additional images around a given image
using concolic testing. \emph{Fuzzing} generates new test data constrained by
an input model and tries to identify interesting test cases, \eg by
optimizing white-box coverage mentioned
above~\cite{odena_tensorfuzz_2019}. Fuzzing techniques may also be
combined with the differential testing approach discussed
above~\cite{guo_dlfuzz_2018}.
In all these cases it needs to be ensured that the image as well as
its meta-data remain valid for testing after transformation.
Finally, \emph{proving methods} surveyed by Liu \etal \cite{Liu_verifying_DNN} try to formally prove properties on a
trained neural network, \eg based on satisfiablity modulo theories (SMT).  
These approaches require a formal characterization of an input space
and the property to be checked, which is hard for non-trivial
properties like contents of an image. 
\par
Existing formal testing approaches can be quite costly to integrate
into testing workflows (\cf \cite{woehrle_open_2019}):
Differential testing and data augmentation require several inferences
per initial test sample;
concolic and fuzzy testing apply an optimization to each given test
sample, while convergence towards the coverage goals is not guaranteed;
also, the iterative approaches need tight integration into the testing
workflow;
and lastly, proving methods usually have to balance computational
efficiency against the precision or completeness of the result
\cite{Liu_verifying_DNN}.
%
Another challenge of formal testing is that machine learning
applications usually solve problems for which no (formal) specification
is possible. This makes it hard to find useful requirements for
testing \cite{zhang:2019} and properties that can be formally
verified. 
Even partial requirements such as specification of useful input
perturbations, specified corner cases, and valuable coverage goals are
typically difficult to identify \cite{schwalbe_survey_2020, Breitenstein2020}.

\subsection{Black Box Methods}\label{subsec:verification:black_box_methods}
\sectionauthor{Jonas Löhdefink\textsuperscript{15}, Julia Rosenzweig\textsuperscript{1}}
In machine learning literature, neural networks are often referred to as black boxes due to the fact that their internal operations and their decision making are not completely understood \cite{OpeningBB2017}, hinting at a lack of interpretability and transparency. However, in this survey we consider a black box to be a machine learning model to which we only have oracle (query) access \cite{Papernot2017,Tramer2016}. That means we can query the model to get input-output pairs, but we do not have access to the specific architecture (or weights, in case of neural networks). As \cite{ReverseEngBB2018} describes, black boxes are increasingly wide-spread, \eg healthcare, autonomous driving or ML as a service in general, due to proprietary, privacy or security reasons.  
\par
As deploying black boxes gains popularity, so do methods that aim to extract internal information such as architecture and parameters or to find out, whether a sample belongs to the training dataset. These include 
\emph{model extraction attacks} \cite{ThievesBERT20,Tramer2016}, \emph{membership inference attacks} \cite{MembershipAttacks2017}, general attempts to reverse-engineer the model \cite{ReverseEngBB2018} or to attack it adversarially \cite{Papernot2017}. Protection and counter-measures are also  actively researched: \cite{ModelExtractionWarning18} proposes a warning system that estimates how much information an attacker could have gained from queries. The authors of \cite{Watermarking2018} use watermarks for models to prevent illegal re-distribution and to identify intellectual property.
\par
Many papers in these fields make use of so-called \emph{surrogate}, \emph{avatar} or \emph{proxy models} that are trained on input-output pairs of the black box.
In case the black-box output is available in soft form (\eg logits), distillation as first proposed by \cite{HintonDistillation15} can be applied to train the surrogate (student) model. Then, any white-box analysis can be performed on the surrogates (\cf \eg \cite{Papernot2017}) to craft adversarial attacks targeted at the black box. More generally, (local) surrogates as for example in \cite{ribeiro_why_2016} can be used to (locally) explain its decision-making. 
Moreover, these techniques are also of interest if one wants to compare or test black-box models (\cf \refsec{subsec:verification:formal_testing}, formal verification). 
This is the case, among others, in ML marketplaces, where you wish to buy a pre-trained model \cite{Watermarking2018}, or if you want to verify or audit that a third-party black-box model obeys  regulatory rules (\cf \cite{Clark2019Regulatory}).

Another topic of active research are so-called \emph{observers}.
The concept of observers is to evaluate the interface of a black-box module to determine if it behaves as expected within a given set of parameters.
The approaches can be divided into \emph{model-explaining} and \emph{anomaly-detecting} observers.
First, model explanation methods answer the question of which input characteristic is responsible for changes at the output.
The observer is able to alter the inputs for this purpose.
If the input of the model under test evolves only slightly but the output changes drastically, this can be a signal that the neural network is mislead, which is also strongly related to adversarial examples (\cf Chapter \ref{sec:adversarial_attacks}).
Hence, the reason for changes in the classification result via the input can be very important.
In order to figure out in which region of an input image the main reason for the classification is located, \cite{Fong_2017} ``delete'' information from the image by replacing regions with generated patches until the output changes.
This replaced region is likely responsible for the decision of the neural network.
Building upon this, \cite{uzunova2019interpretable} adapt the approach to medical images and generate ``deleted'' regions by a variational autoencoder (VAE).
Second, anomaly-detecting observers register input and output anomalies, either examining input and output independently or as an input-output pair, and predict the black-box performance in the current situation.
In contrast to model-explaining approaches, this set of approaches has high potential to be used in an online scenario since it does not need to modify the model input.
The maximum mean discrepancy (MMD)~\cite{borgwardt2006integrating} measures the domain gap between two data distributions independently from the application and can be used to raise a warning if input or output distributions during inference deviate too strongly from their respective training distributions.
By use of a GAN-based autoencoder \cite{loehdefink2020} perform a domain shift estimation using neural networks in conjunction with the Wasserstein distance as domain mismatch metric.
This metric can also be evaluated by use of a casual time-variant aggregation of distributions during inference time.

\section{Architecture}\label{sec:architecture}
\sectionauthor{Michael Weber\textsuperscript{14}}
In order to solve a specific task, the architecture of a CNN and its building blocks play a significant role.
Since the early days of using CNNs in image processing, when they were applied to handwriting recognition~\cite{lecun89} and the later breakthrough in general image classification~\cite{Krizhevsky2012}, the architecture of the networks has changed radically.
Did the term of \emph{deep learning} for these first convolutional neural networks imply a depth of approximately four layers, their depth increased significantly during the last years and new techniques had to be developed to successfully train and utilize these networks~\cite{he16}.
In this context, new activation functions~\cite{ramachandran18} as well as new loss functions~\cite{lin17a} have been designed and new optimization algorithms~\cite{kingma15} were investigated.

With regard to the layer architecture, the initially alternating repetition of convolution and pooling layers as well as their characteristics have changed significantly.
The convolution layers made the transition from a few layers with often large filters to many layers with small filters.
A further trend was then the definition of entire modules, which were used repeatedly within the overall architecture as so-called \emph{network in network}~\cite{lin14}.

In areas such as autonomous driving, there is also a strong interest in the simultaneous execution of different tasks within one single convolutional neural network architecture.
This kind of architecture is called \emph{multi-task learning}~(MTL)~\cite{Caruana1997} and can be utilized in order to save computational resources and at the same time to increase performance of each task \cite{Klingner2020a}.
Within such multi-task networks, usually one shared feature extraction part is followed by one separate so-called head per task~\cite{Teichmann2018}.

In each of these architectures, manual design using expert knowledge plays a major role. The role of the expert is the crucial point here.
In recent years, however, there have also been great efforts to automate the process of finding architectures for networks or, in the best case, to learn them.
This is known under the name \emph{neural architecture search}~(NAS).

\subsection{Building Blocks}\label{subsec:architecture:building_blocks}
\sectionauthor{Michael Weber\textsuperscript{14}}
Designing a convolutional neural network typically includes a number of design choices.
The general architecture usually contains a number of convolutional and pooling layers which are arranged in a certain pattern. 
Convolutional layers are commonly followed by a non-linear activation function. 
The learning process is based on a loss function which determines the current error and an optimization function that propagates the error back to the single convolution layers and its learnable parameters.

When CNNs became state of the art in computer vision~\cite{Krizhevsky2012}, they were usually built using a few alternating convolutional and pooling layers having a few fully connected layers in the end.
It turned out that better results are achieved with deeper networks and so the number of layers increased~\cite{simonyan14} over the years.
To deal with these deeper networks, new architectures had to be developed.
In a first step, to reduce the number of parameters, the convolutional layers with partly large filter kernels were replaced by several layers with small $3 \times 3$ kernels.
Today, most architectures are based on the \emph{network in network} principle~\cite{lin14}, where more complex modules are used repeatedly. 
Examples of such modules are the \emph{inception module} from GoogleNet~\cite{szegedy_going_2015} or the \emph{residual block} from ResNet~\cite{he16}. 
While the inception module consists of multiple parallel strings of layers, the residual blocks are based on the \emph{highway network}~\cite{srivastava15}, which means that they can bypass the original information and the layers in between are just learning residuals.
With ResNeXt~\cite{xie17} and Inception-ResNet~\cite{szegedy17} there already exist two networks that combine both approaches.
For most tasks, it turned out that replacing the fully connected layers by convolutional layers is much more convenient making the networks fully convolutional~\cite{long15}.
These so-called \emph{fully convolutional networks}~(FCN) are no longer bound to fixed input dimensions.
Note that with the availability of convolutional long short-term memory (ConvLSTM) structures also fully convolutional recurrent neural networks (FCRNs) became available for fully scalable sequence-based tasks \cite{Shi2015,Strake2020}.

Inside the CNNs, the \emph{rectified linear unit}~(ReLU) has been the most frequently used activation function for a long time.
However, since this function suffers from problems related to the mapping of all negative values to zero like the vanishing gradient problem, new functions have been introduced in recent years.
Examples are the \emph{exponential linear unit}~(ELU), \emph{swish}~\cite{ramachandran18} and the \emph{non-parametric linearly scaled hyperbolic tangent}~(LiSHT)~\cite{roy19}.
In order to be able to train a network consisting of these different building blocks, 
the loss function is the most crucial part. This function is responsible for how and what the network ultimately learns and how exactly the training data is applied during the training process to make the network train faster or perform better. So the different classes can be weighted in a classification network with fixed values or so-called $\alpha$-balancing according to their probability of occurrence. Another interesting approach is weighting training examples according to their easiness for the current network~\cite{lin17a},~\cite{weber19}. For multi-task learning also weighting tasks based on their uncertainty~\cite{Kendall2018} or gradients~\cite{Chen2018} can be done as further explained in Sec.~\ref{subsec:architecture:multi_task_networks}.
A closer look on how a modification of the loss function might affect safety-related aspects is given in Sec.~\ref{subsec:robust_training:modification_of_loss}.

\subsection{Multi-Task Networks}\label{subsec:architecture:multi_task_networks}
\sectionauthor{Marvin Klingner\textsuperscript{15}, Varun Ravi-Kumar\textsuperscript{4}, Timo Sämann\textsuperscript{4}, Gesina Schwalbe\textsuperscript{3}}
\emph{Multi-task learning} (MTL) in the context of neural networks describes the process of optimizing several tasks simultaneously by learning a unified feature representation \cite{Caruana1997, Guizilini2020, Rebuffi2018, Klingner2020} and coupling the task-specific loss contributions, thereby enforcing cross-task consistency \cite{Casser2019, Luo2019a, Klingner2020a}.\par
Unified feature representation is usually implemented by sharing the parameters of the initial layers inside the encoder (also called feature extractor). It not only improves the single tasks by more generalized learned features but also reduces the demand for computational resources at inference. Not an entirely new network has to be added for each task but only a task-specific decoder head. It is essential to consider the growing amount of visual perception tasks in autonomous driving, e.g., depth estimation, semantic segmentation, motion segmentation, and object detection. While the parameter sharing can be soft, as in \emph{cross stitch} \cite{Misra2016} and \emph{sluice networks} \cite{Ruder2017}, or hard \cite{Teichmann2018, Kokkinos2017}, meaning ultimately sharing the parameters, the latter is usually preferred due to its straightforward implementation and lower computational complexity during training and inference.\par
Compared to implicitly coupling tasks via a shared feature representation, there are often more direct ways to optimize the tasks inside cross-task losses jointly. It is only made possible as, during MTL, there are network predictions for several tasks, which can be enforced to be consistent. As an example, sharp depth edges should only be at class boundaries of semantic segmentation predictions. Often both approaches to MTL are applied simultaneously~\cite{Chen2019, Yang2018} to improve a neural network's performance as well as to reduce its computational complexity at inference.\par
While the theoretical expectations for MTL are quite clear, it is often challenging to find a good weighting strategy for all the different loss contributions as there is no theoretical basis on which one could choose such a weighting with early approaches either involving heuristics or extensive hyperparameter tuning. The easiest way to balance the tasks is to use uniform weight across all tasks. However, the losses from different tasks usually have different scales, and uniformly averaging them suppresses the gradient from tasks with smaller losses. Addressing these problems, Kendall et al. \cite{Kendall2018} propose to weigh the loss functions by the \emph{homoscedastic uncertainty} of each task. One does not need to tune the weighting parameters of the loss functions by hand, but they are adapted automatically during the training process. Concurrently Chen et al.~\cite{Chen2018} propose \emph{GradNorm}, which does not explicitly weigh the loss functions of different tasks but automatically adapts the gradient magnitudes coming from the task-specific network parts on the backward pass. Liu et al. \cite{liu2019end} proposed dynamic weight average (DWA), which uses an average of task losses over time to weigh the task losses.
\subsection{Neural Architecture Search} \label{subsec:architecture:neural_architecture_search}
\sectionauthor{Patrick Feifel\textsuperscript{8}, Seyed Eghbal Ghobadi\textsuperscript{8}}
In the previous sections we saw manually engineered modifications of existing CNN architectures proposed by ResNet \cite{he16} or Inception \cite{szegedy_going_2015}. They are results of human design and showed their ability to improve performance. ResNet introduces a \emph{skip connection} in building blocks and Inception makes use of its specific \emph{inception module}. Hereby, the intervention by an expert is crucial. The approach of \emph{neural architecture search} (NAS) aims to automate this time-consuming and manual design of neural network architectures. 
\par
NAS is closely related to hyperparameter optimization (HO), which is described in \refsec{subsec:robust_training:hyperparameter_optimization}. Originally, both tasks were solved simultaneously. Consequently, the kernel size or number of filters were seen as additional hyperparamters. Nowadays, the distinction between HO and NAS should be stressed. The concatenation of complex building blocks or modules cannot be accurately described with single parameters. This simplification is no longer suitable. 
\par
To describe the NAS process, the authors of \cite{elsken_neural_2019} define three steps: (1) definition of search space, (2) search strategy and (3) performance estimation strategy. 
\par
The majority of search strategies take advantage of the \emph{NASNet search space} \cite{zoph_learning_2018} which arranges various operations, \eg convolution, pooling within a single cell. However, other spaces based on a chain or multi-branch structure are possible \cite{elsken_neural_2019}. The search strategy comprises advanced methods from 
sequential model-based optimization (SMBO) \cite{liu_progressive_2018},
Bayesian optimization \cite{kandasamy_neural_2018}, 
evolutionary algorithms \cite{real_regularized_2019, elsken_efficient_2019}, 
reinforcement learning \cite{zoph_learning_2018, pham_efficient_2018} and 
gradient descent \cite{liu_darts_2019, stamoulis_single-path_2019}. 
Finally, the performance estimation describes approximation techniques due to the impracticability of multiple evaluation runs. For a comprehensive survey regarding the NAS process we refer to \cite{elsken_neural_2019}. 
\par
Recent research has shown that reinforcement learning approaches such as NASNet-A \cite{zoph_learning_2018} and ENAS \cite{pham_efficient_2018} are partly outperformed by evolutionary algorithms, \eg AmoebaNet \cite{real_regularized_2019} and gradient-based approaches, \eg DARTS \cite{liu_darts_2019}. 
\par
Each of these approaches focuses on different optimization aspects. Gradient-based methods are applied to a continuous search space and offer faster optimization. On the contrary, the evolutionary approach LEMONADE \cite{elsken_efficient_2019} enables multi-object optimization by considering the conjunction of resource consumption and performance as the two main objectives. Furthermore, single-path NAS \cite{stamoulis_single-path_2019} extends the multi-path approach of former gradient-based methods and proposes the integration of 'over-parameterized superkernels', which significantly reduces memory consumption.
\par
The focus of NAS is on the optimized combination of humanly predefined CNN elements with respect to objectives such as resource consumption and performance. NAS offers automation, however, the realization of the objectives is strongly limited by the potential of the CNN elements. 
\section{Model Compression}\label{sec:compression}
\sectionauthor{Serin Varghese\textsuperscript{7}}
Recent developments in CNNs have resulted in neural networks being the state-of-the-art in computer vision tasks like image classification~\cite{Krizhevsky2012, He2015, Mahajan2018}, object detection~\cite{Girshick2015, Redmon2015, He2017} and semantic segmentation~\cite{Chen2017, Zhu2019, Wang2019, Loehdefink2019}. This is largely due to the increasing availability of hardware computational power and an increasing amount of training data. We also observe a general upwards trend of the complexities of the neural networks along with their improvement in state-of-the-art performance. These CNNs are largely trained on back-end servers with significantly higher computing capabilities. The use of these CNNs in real-time applications are inhibited due to the restrictions on hardware, model size, inference time, and energy consumption. 
This led to an emergence of a new field in machine learning, commonly termed as model compression. Model compression basically implies reducing the memory requirements, inference times and model size of DNNs to eventually enable the use of neural networks on edge devices. This is tackled by different approaches such as \emph{network pruning} (identifying weights or filters that are not critical for network performance), \emph{weight quantizations} (reducing the precision of the weights used in the network), \emph{knowledge distillation} (a smaller network is trained with the knowledge gained by a bigger network), and \emph{low-rank factorization} (decomposing a tensor into multiple smaller tensors).
In this section, we introduce some of these methods for model compression and discuss in brief the current open challenges and possible research directions with respect to its use in automated driving applications.

\subsection{Pruning}\label{subsec:compression:pruning}
\sectionauthor{Falk Kappel\textsuperscript{13}, Serin Varghese\textsuperscript{7}}
Pruning has been used as a systematic tool to reduce the complexity of deep neural networks. The redundancy in DNNs may exist on various levels, such as the individual weights, filters, and even layers. All the different methods for pruning try to take advantage of these available redundancies on various levels. Two of the initial approaches for neural networks proposed weight pruning in the 1990s as a way of systematically damaging neural networks~\cite{Cun1990, Reed1993}. As these weight pruning approaches do not aim at changing the structure of the neural network, these approaches are called \emph{unstructured pruning}. Although there is reduction in the size of the network when it is saved in sparse format, the acceleration depends on the availability of hardware that facilitate sparse multiplications. As pruning filters and complete layers aim at exploiting the available redundancy in the architecture or structure of neural networks, these pruning approaches are called \emph{structured pruning}. 
Pruning approaches can also be broadly classified into: data-dependent and data-independent methods. Data-dependent methods~\cite{Liu2017, Luo2017, He2017a} make use of the training data to identify filters to prune. Theis \etal \cite{Theis2018} and Molchanov \etal \cite{Molchanov2017} propose a greedy pruning strategy that identifies the importance of feature maps one at a time from the network and measures the effect of removal of the filters on the training loss. This means that filters corresponding to those feature maps that have least effect on training loss are removed from the network. Within data-independent methods~\cite{Li2017, He2018, Ye2018, Zhuo2018}, the selection of CNN filters to be pruned are based on the statistics of the filter values. Li \etal \cite{Li2017} proposed a straightforward method to calculate the rank of filters in a CNN. The selection of filters are based on the $\ell_1$-norm, where the filter with the lowest norm is pruned away. He \etal \cite{He2017a} employ a LASSO regression-based selection of filters to minimize the least squares reconstruction.\par
Although the above-mentioned approaches demonstrated that a neural network can be compressed without affecting the accuracy, the effect on robustness is largely unstudied.  Dhillon \etal \cite{Dhillon2018} proposed pruning a subset of activations and scaling up the survivors to show improved adversarial robustness of a network. Lin \etal \cite{Lin2019} quantize the precision of the weights after controlling the Lipschitz constant of layers. This restricts the error propagation property of adversarial pertubations within the neural network. Ye \etal \cite{Ye2019} evaluated the relationship between adversarial robustness and model compression in detail and show that naive compression has a negative effect on robustness. Gui \etal \cite{Gui2019} co-optimize robustness and compression constraints during the training phase and demonstrate improvement in the robustness along with reduction in the model size. However, these approaches have mostly been tested on image classification tasks and on smaller datasets only. Their effectiveness on safety-relevant automated driving tasks such as object detection and semantic segmentation tasks are not studied and remains an open research challenge. 
\subsection{Quantization}\label{subsec:compression:quantization}
\sectionauthor{Firas Mualla\textsuperscript{13}}
Quantization of a random variable $x$ having a probability density function $f(x)$ is the process of dividing the range of $x$ into intervals, each is represented using a single value (also called reconstruction value), such that the following reconstruction error is minimized:
\begin{equation}
\label{eq:quantization}
\sum_{i=1}^{L} \int_{b_i}^{b_{i+1}} (q_i - x)^2 f(x) dx,
\end{equation}
where $b_i$ is the left-side border of the i-th interval, $q_i$ is its reconstruction value, and $L$ is the number of intervals, \eg $L = 8$ for a 3-bit quantization. This definition can be extended to multiple dimensions as well.

Quantization of neural networks has been around since the 1990s \cite{guo2018:survey}, however, with a focus in the early days on improving the hardware implementations of these networks. In the deep learning literature, a remarkable application of quantization combined with unstructured pruning can be found in the approach of \textit{deep compression} \cite{han2016:deepcompression}, where 1-dimensional k-means is utilized to cluster the weights per layer and thus finding the $L$ cluster centers ($q_i$ values in (\ref{eq:quantization})) iteratively. This procedure conforms to an implicit assumption that $f(x)$ has the same spread inside all clusters. Deep compression can reduce the network size needed for image classification by a factor of 35 for AlexNet and a factor of 49 for VGG-16 without any loss in accuracy. However, as pointed out in \cite{Jacob2018:integeronly}, these networks from the early deep learning days are over-parameterized and a less impressing compression factor is thus expected when the same technique is applied to lightweight architectures such as MobileNet and SequeezeNet. For instance, considering SqueezeNet (50 times smaller than AlexNet), the compression factor of deep compression without accuracy loss drops to about 10.

Compared to scalar quantization used in deep compression, there were attempts to exploit the structural information by applying variants of vector quantization of the weights \cite{Gong2014:VQ,Choi2020:universal,stock2020:bitdown}. Remarkably, in the latter (\ie \cite{stock2020:bitdown}), the reconstruction error of the activations (instead of the weights) is minimized in order to find an optimal codebook for the weights, as the ultimate goal of quantization is to approximate the network's output not the network itself. This is performed in a layer-by-layer fashion (as to prevent error accumulation) using activations generated from unlabeled data.

Other techniques \cite{Mathew2017:tidl, Jacob2018:integeronly} apply variants of so-called ``linear'' \textit{quantization}, \ie the quantization staircase has a fixed interval size. This paradigm conforms to an implicit assumption that $f(x)$ in (\ref{eq:quantization}) is uniform and is thus also called \textit{uniform quantization}. The uniform quantization is widely applied both in specialized software packages such as the \texttt{Texas Instruments Deep Learning Library} (automotive boards) \cite{Mathew2018:tidlwp} and in general-purpose libraries such as the \texttt{Tensorflow Lite}. The linearity assumption enables practical implementations, as the quantization and dequantization can be implemented using a scaling factor and an intercept, whereas no codebook needs to be stored. In many situations, the intercept can be omitted by employing a symmetric quantization mapping. Moreover, for power of 2 ranges, the scaling ends up being a bitwise shift operator, where quantization and dequantization differ only in the shift direction. It is also straightforward to apply this scheme \textit{dynamically}, \ie for each tensor separately using a tensor-specific multiplicative factor. This can be easily applied not only to filters (weight tensors) but also to activation tensors (see for instance \cite{Mathew2017:tidl}).

Unless the scale factor in the linear quantization is assumed constant by construction, it is computed based on the statistics of the relevant tensor and can be thus sensitive to outliers. This is known to result in a low precision quantization. In order to mitigate this issue, the original range can be \textit{clipped} and thus reduced to the most relevant part of the signal. Several approaches are proposed in the literature for finding an optimal clipping threshold: simple percentile analysis of the original range (\eg clipping 2\% of the largest magnitude values), minimizing the mean square error between the quantized and original range in the spirit of (\ref{eq:quantization}) \cite{Banner2019:mseclip}, or minimizing the Kullback-Leibler divergence between the original and the quantized distributions \cite{migacz2017:8BitTensorRT}. While the clipping methods trade off large quantization errors of outliers against small errors of inliers \cite{wu2020:IntQuantPrinciples}, other methods tackle the outliers problem using a different trade-off, see for instance the outlier channel splitting approach in \cite{zhao2019:OCS}.

An essential point to consider when deciding for a quantization approach for a given problem is the allowed or intended interaction with the training procedure. The so-called \textit{post-training quantization}, \ie quantization of a pre-trained network, seems to be attractive from a practical point of view: No access to training data is required and the quantization and training toolsets can be independent from each other. On the other hand, the training-aware quantization methods often yield higher inference accuracy and shorter training times. The latter is a serious issue for large complicated models which may need weeks to train on modern GPU clusters.
The training-aware quantization can be implemented by inserting fake quantization operators in the computational graph of the forward-pass during training (simulated quantization), whereas the backward pass is done as usual in floating-point resolution \cite{Jacob2018:integeronly}. Other approaches \cite{Zhu2019:UnifiedInt8,zhou2016:dorefa} go a step further by quantizing the gradients as well. This leads to much lower training time, as the time of the often computationally expensive backward pass is reduced. The gradient's quantization, however, is not directly applicable as it requires the derivative of the quantization function (staircase-like), which is zero almost everywhere. Luckily, this issue can be handled by employing a \textit{straight-through estimator} \cite{bengio2013:ste} (approximating the quantization function by an identity mapping). There are also other techniques proposed recently to mitigate this problem \cite{Uhlich2019:DiffQuant, Liu2019:WithoutSTE}.

\section{Discussion}\label{sec:discussion}
We have presented an extensive overview of approaches to effectively handle safety concerns accompanying deep learning: lack of generalization, robustness, explainability, plausibility, and efficiency. 
It has been described which lines of research we deem prevalent, important, and promising for each of the individual topics and categories into which the presented methods fall. 
\par
The reviewed methods alone will not provide safe ML systems as such -- and neither will their future extensions. This is due to the limitations of quantifying complex real-world contexts. A complete and plausible \emph{safety argumentation} will, thus, require more than advances in methodology and theoretical understanding of neural network properties and training processes. 
Apart from methodological progress, it will be necessary to gain practical experience in using the presented methods to gather evidence for overall secure behavior, using this evidence to construct a tight safety argument, and testing its validity in various situations.
\par
In particular, each autonomously acting robotic system with state-of-the-art deep-learning-based perception and non-negligible actuation may serve as an object of study and is, in fact, in need of this kind of systematic reasoning before being transferred to widespread use or even market entry. We strongly believe that novel scientific insights, the potential market volume, and public interest will drive the arrival of reliant and trustworthy AI technology.

\section*{Acknowledgment}
The research leading to these results is funded by the German Federal Ministry for Economic Affairs and Energy within the project “KI Absicherung – Safe AI for Automated Driving”. The authors would like to thank the consortium for the successful cooperation. Furthermore, this research has been funded by the Federal Ministry of Education and Research of Germany as part of the competence center for machine learning ML2R (01IS18038B).

\bibliography{main} 
\bibliographystyle{alpha}

\end{document}